\documentclass{article}

\PassOptionsToPackage{numbers, compress}{natbib}

\usepackage[preprint]{neurips_2023}

\usepackage[utf8]{inputenc} % allow utf-8 input
\usepackage[T1]{fontenc}    % use 8-bit T1 fonts
\usepackage{hyperref}       % hyperlinks
\usepackage{url}            % simple URL typesetting
\usepackage{booktabs}       % professional-quality tables
\usepackage{amsfonts}       % blackboard math symbols
\usepackage{nicefrac}       % compact symbols for 1/2, etc.
\usepackage{microtype}      % microtypography
\usepackage{xcolor}         % colors
\usepackage{algorithm}
\usepackage{algorithmic}
\usepackage{amsmath}
\usepackage{graphicx}
\usepackage{enumitem}

\title{Insufficiently Justified Disparate Impact: A New Criterion for Subgroup Fairness}

\author{%
  Neil Menghani \\
  New York University \\
  New York, NY \\
  \texttt{nlm326@nyu.edu} \\
  \And
  Edward McFowland III \\
  Harvard Business School \\
  Boston, MA \\
  \texttt{emcfowland@hbs.edu} \\
  \AND
  Daniel B. Neill \\
  New York University \\
  New York, NY \\
  \texttt{daniel.neill@nyu.edu} \\
}

\begin{document}

\maketitle

\begin{abstract}
  In this paper, we  develop a new criterion, ``insufficiently justified disparate impact'' (IJDI), for assessing whether recommendations (binarized predictions) made by an algorithmic decision support tool are fair. Our novel, utility-based IJDI criterion evaluates false positive and false negative error rate imbalances, identifying statistically significant disparities between groups which are present even when adjusting for group-level differences in base rates. We describe a novel IJDI-Scan approach which can efficiently identify the intersectional subpopulations, defined across multiple observed attributes of the data, with the most significant IJDI. To evaluate IJDI-Scan’s performance, we conduct experiments on both simulated and real-world data, including recidivism risk assessment and credit scoring. Further, we implement and evaluate approaches to mitigating IJDI for the detected subpopulations in these domains.
\end{abstract}

\section{Introduction} \label{sec:intro}

Public sector decision-makers often have implicit biases due to their own assumptions and exposure to a limited sample of data, thus leading to disparate outcomes. For example, policing practices and judges’ court decisions have been found to demonstrate racial biases~\cite{gelman07,corbett-davies17}. The presence of bias in human decision-making suggests the potential for algorithmic decision support tools to improve the quality and fairness of these decisions. However, there are many instances in which algorithms can either exacerbate existing biases or create new ones~\cite{angwin2016machine, bolukbasi16, buolamwini18}.

Here we focus on the common setting in which an algorithmic decision support tool provides binary \emph{recommendations} to inform decisions or actions, based on its probabilistic \emph{predictions} of the likelihood that an event will occur. For example, an algorithm may estimate loan applicants' risk of default and only recommend loans to lower-risk individuals.  In criminal justice, the COMPAS risk assessment tool has been used to predict defendants' probability of reoffending and classify each defendant as ``high-'' or ``low-risk'', greatly impacting outcomes such as bail, sentencing, and parole. 

For these high-stakes decisions, \emph{error rate balance}~\cite{hardt2016,mehrabi2021survey} is a commonly used fairness criterion, since errors can have substantial impacts on individuals' lives. For example, a ProPublica study found that COMPAS exhibited higher false positive rate (FPR) and lower false negative rate (FNR) when evaluating Black defendants~\cite{angwin2016machine}, thus 
leading to disparate outcomes harming Black defendants.

In this paper, we explicitly assess bias in (binarized) recommendations rather than (probabilistic) predictions.  Our motivation is that error rate imbalances can be the result of how predictive models are used in practice, not necessarily the prediction instruments themselves.  For example, given a perfectly calibrated risk predictor, it has been proven that using a fixed threshold for determining ``high'' vs. ``low'' risk leads to error rate imbalances whenever base rates differ between groups, while allowing group-dependent thresholds could mitigate these imbalances~\cite{chouldechova2017fair}.

Here we consider the question of how to assess error rate imbalances while also accounting for differences in base rates (e.g., true probability of reoffending) between groups.  As Corbett-Davies et al.~\cite{corbett2018measure} state, 
``because the risk distributions of protected groups will in general differ, threshold-based decisions will typically yield error metrics that also differ by group\ldots higher FPR for one group relative to another may either mean that the higher FPR group faces a lower threshold, indicating discrimination, or\ldots has a higher base rate.''  This motivates us to ask, ``How much difference in error rates should be allowed before we flag this difference as evidence of discrimination?''  

One approach~\cite{jung2019} is to only compare error rates across groups after fully controlling for differences in base rates. This \emph{risk-adjusted regression} criterion allows for FPR to be any monotonic function of base rate that is independent of group membership and still be considered ``fair'': any disparate impacts are \emph{justified} by the difference in base rates between groups.  However, this approach can lead to huge disparities between groups.  Consider an extreme example where we have Group A of defendants who each have 51\% probability of reoffending and Group B of defendants who each have 49\%. Assume these probabilities are known and compared to a fixed threshold to classify ``high-risk'' versus ``low-risk'' defendants.  Here the probabilistic risk predictions are equal to each individual's true probability of reoffending, and thus perfectly calibrated (fair, in the sense of not being systematically biased upward or downward from their true values).  However, given a threshold of 50\%, Group A would have a 100\% false positive rate versus Group B’s 0\%.  This huge disparity would be considered ``justified'' by~\cite{jung2019}, but we argue that it would not be \emph{sufficiently} justified by the small difference in base rates. However, this result does not negate the usefulness of considering base rates when comparing across groups. If Group A all had a 99\% probability of reoffending and Group B only 1\%, we could consider this huge difference in base rates as a sufficient reason for treating the two groups differently, thus justifying the error rate disparity.  Moreover, requiring near-exact balance ($\epsilon$-fairness) in error rates, regardless of base rate differences, can lead to suboptimal solutions from the perspective of welfare maximization~\cite{hu2020}.  This motivates our approach of defining the ``allowable'' difference in error rates between groups as a function of their difference in base rates, using a utility-based formulation to assess the tradeoff between error rate balance and social welfare.

Thus we propose a new criterion, \emph{insufficiently justified disparate impact}, or IJDI, that considers differences in base rates as well as error rate disparities when evaluating whether recommendations are fair across groups.  We derive our IJDI definitions from a general notion of utility, under the assumption that error rate imbalances are undesirable (creating a disutility proportional to the amount of imbalance), and then assess whether differences in utility are large enough to sufficiently justify a disparity in error rates. Our work makes the following contributions:

\begin{itemize}[noitemsep,topsep=0pt,parsep=0pt,partopsep=0pt,leftmargin=5mm] 
\item We derive a new utility-based IJDI fairness criterion to evaluate the fairness of recommendations.
\item We create a novel algorithmic approach, IJDI-Scan, 
to search over intersectional subpopulations and identify the subpopulations with the most significant violations of the IJDI fairness criterion.
\item IJDI-Scan is a further generalization of two new extensions of Bias Scan~\cite{biasscan}, the FPR- and TPR-Scans, which can detect significant error rate imbalances but do not incorporate utility. 
\item While FPR-Scan and TPR-Scan can be reduced to the original Bias Scan by preprocessing, generalizing these scans to detect IJDI requires multiple, non-trivial algorithmic extensions.
\item We demonstrate the effectiveness of the IJDI-Scan approach for detecting subgroups violating the IJDI criterion by performing experiments on simulated and real-world data.
\item We explore approaches for mitigating IJDI to improve the fairness of recommendations and implement the best approaches on data across multiple domains.
\end{itemize}

\section{Background: Bias Scan}

Given a dataset $D$ with $M$ categorical features, let us consider an intersectional \emph{subgroup} to be defined by $M$ non-empty subsets of attribute values, one for each feature $X_m$ of $D$. Thus there are $\prod_{m=1\ldots M}(2^{|X_m|}-1)$ possible subgroups.

Our objective in evaluating recommendations is to determine the subgroups which most significantly violate the new IJDI criterion we define below. This is a computationally challenging task, as there are exponentially many subgroups over which to scan. However, Bias Scan~\cite{biasscan} can be used to perform conditional optimization over the $2^{|X_m|}-1$ possible non-empty subsets for a given attribute $X_m$, conditional on the current subsets of values for all other attributes, in linear rather than exponential time.  The scan iterates over each attribute until it arrives at a local maximum of its score function (a log-likelihood ratio statistic), and uses multiple random restarts to approach the global maximum. Statistical significance of the highest-scoring subgroups can be obtained by randomization testing.

The traditional Bias Scan approach looks for \emph{miscalibration bias} in predictions, comparing the observed binary outcomes $y_i$ to the predicted probabilities, $\hat{p}_i$, that $y_i=1$, and identifying intersectional subgroups $S$ where the probabilities $\hat p_i$ are systematically biased upward or downward.  The Bias Scan score function is defined as a log-likelihood ratio between two hypotheses. The null hypothesis is that the predictions $\hat{p}_i$ represent the correct odds of $y_i = 1$ for all $i$, while the alternative hypothesis is that $\hat{p}_i$ over- or under-estimates the odds by some multiplicative factor $q$ in subgroup $S$.
These hypotheses can be formulated as $H_0: odds(y_i = 1) = \frac{\hat{p}_i}{1-\hat{p}_i}, \forall i \in D$ and $H_1(S): odds(y_i = 1) = \frac{q\hat{p}_i}{1-\hat{p}_i}, \forall i \in S;$
$odds(y_i = 1) = \frac{\hat{p}_i}{1-\hat{p}_i}, \forall i \in D\setminus S,$
where $q$ is obtained by maximum likelihood, subject to constraints $q>1$ to detect underestimated probabilities or $0<q<1$ to detect overestimated probabilities respectively. The corresponding log-likelihood ratio score is $F(S) = \max_q \left( \sum_{i \in S} y_i \log (q) - \sum_{i \in S} \log (1 - \hat{p}_i + q\hat{p}_i) \right)$.

\section{A New Extension of Bias Scan for Error Rate Balance} \label{extension}

As an initial step toward our IJDI criterion, we extend the traditional Bias Scan to evaluate other measures of fairness beyond calibration of predictions, specifically focusing on error rate balance both for false positive rates and false negative rates.  To do so, we use the original Bias Scan as a building block, but redefine the binary outcomes $y_i$ and probabilistic predictions $\hat{p}_i$ for the scan, which we now denote as $y_{i,scan}$ and $\hat{p}_{i, scan}$ respectively.

For each individual $i$, let $y_{i,0} \in \{0,1\}$ denote the original binary outcome from the dataset, $\hat{p}_{i,0} \in [0,1]$ denote the classifier's probabilistic prediction of $\Pr(y_{i,0} = 1)$, and $\hat{p}_{i,b} \in \{0,1\}$ denote the binary recommendation corresponding to that prediction.  For the COMPAS data, $y_{i,0}$ represents whether or not individual $i$ reoffended, $\hat{p}_{i,0}$ represents COMPAS's estimated probability that individual $i$ would reoffend (derived from the decile score by maximum likelihood), and $\hat{p}_{i,b}$ represents the recommendation to treat the individual as ``high risk'' or ``low risk'' based on the COMPAS prediction.  Typically, $\hat{p}_{i,b}$ is calculated by comparing $\hat{p}_{i,0}$ to a fixed threshold $\theta$, i.e., $\hat{p}_{i,b}=1$
if $\hat{p}_{i,0}\ge\theta$, and $\hat{p}_{i,b}=0$ otherwise.
 
Our false positive rate scan (FPR-Scan) identifies subgroups $S$ with a higher proportion of false positive errors, $\hat{p}_{i,b}=1$ for individuals with $y_{i,0} = 0$, as compared to the rest of the population. To do so, we first filter the dataset to only those individuals with $y_{i,0} = 0$ (``negatives''); denote the resulting dataset by $D_N$.  Then $\hat{p}_{i,b}$ represents the outcome of interest, a false positive error for individual $i$, and thus we can replace 
$y_i$ in the traditional Bias Scan by setting $y_{i, scan}=\hat{p}_{i,b}$.  Our null hypothesis is that the FPR is equal for all individuals, and thus we can replace $\hat{p}_i$ in the traditional Bias Scan by setting $\hat{p}_{i, scan} = \bar{p}_b$ for all $i \in D_N$, where $\bar{p}_b = \frac{1}{|D_N|} \sum_{i \in D_N} \hat{p}_{i,b}$ is the average false positive rate across all negative individuals.  We then pass the $y_{i, scan}$ and $\hat{p}_{i, scan}$ to Bias Scan, so that we are testing the hypotheses:
\begin{gather}
    \label{bias-scan-hypothesis}
    \begin{split}
        H_0: odds(\hat{p}_{i,b} = 1) &= \frac{\bar{p}_b}{1-\bar{p}_b}, \forall i \in D_N, \\
        H_1(S): odds(\hat{p}_{i,b} = 1) &= \frac{q\bar{p}_b}{1-\bar{p}_b}, \forall i \in S_N, \\
        odds(\hat{p}_{i,b} = 1) &= \frac{\bar{p}_b}{1-\bar{p}_b}, \forall i \in D_N \setminus S_N,
    \end{split}
\end{gather}
where $S_N$ represents the negative individuals in subgroup $S$, and we constrain $q>1$ to search for subgroups with significantly increased false positive rate.

Similarly, we can define the true positive rate scan (TPR-Scan) to identify subgroups $S$ with a higher proportion of true positives, or equivalently a lower proportion of false negative errors.  To do so, we first filter the dataset to only those individuals with $y_{i,0} = 1$ (``positives''); denote the resulting dataset by $D_P$. Then $\hat{p}_{i,b}$ represents the outcome of interest, a true positive for individual $i$, and thus we set $y_{i, scan}=\hat{p}_{i,b}$.  Our null hypothesis is that the TPR is equal for all individuals, and thus we set $\hat{p}_{i, scan} = \bar{p}_b$ for all $i \in D_P$, where $\bar{p}_b$ is the average TPR across all positive individuals, $\bar{p}_b = \frac{1}{|D_P|} \sum_{i \in D_P} \hat{p}_{i,b}$.  We then pass the $y_{i, scan}$ and $\hat{p}_{i, scan}$ to Bias Scan, and constrain $q>1$ to search for subgroups $S$ where the positive individuals $S_P$ have significantly increased TPR.

The  benefit of formulating this scan using true positive rate rather than false negative rate is that the hypotheses being tested~(eqn.~\eqref{bias-scan-hypothesis}), and therefore the implementations of the FPR- and TPR-Scan, are essentially identical. Both scans search for subgroups with a higher proportion of test positives, $\hat{p}_{i,b}=1$, within different subsets of the data.

\section{Insufficiently Justified Disparate Impact: Fairness Criteria}
\label{sec:ijdi}

The FPR- and TPR-Scans define ``fairness'' as equal FPR and TPR for all individuals and identify intersectional subgroups with the most significant violations of this fairness criterion.  Thus these scans fail to account for base rate differences when assessing error rate imbalances: as discussed in Section~\ref{sec:intro}, we should not expect identical error rates when base rate differences are large.  In this section, we focus on the subpopulation $D_N$ of negative individuals ($y_{i,0} = 0$), and thus on identifying subgroups with significant FPR disparities after adjusting for base rate differences. TPR disparities for the subpopulation $D_P$ of positive individuals can be similarly adjusted, as shown in Appendix~\ref{sec:lambda-p}.  

For any subgroup $S$, let 
$p(S) = \frac{1}{|S_N|} \sum_{i \in S_N} p_i$, and $p({{\sim}} S) = \frac{1}{|D_N \setminus S_N|} \sum_{i \in D_N \setminus S_N} p_i$, where $p_i$ is the true probability $\Pr(y_i = 1 \:|\: x_i)$. In practice, $p_i$ is often not known, but can be estimated from dataset $D$ by any sufficiently rich class of models. Note that this estimate of $p_i$ is distinct from $\hat p_{i,0}$, the estimate of $\Pr(y_i = 1)$ made by the classifier that we are auditing.  For example, for COMPAS data, $\hat p_{i,0}$ is the COMPAS probability estimate, which uses different features to predict reoffending risk, and was learned using data from different jurisdictions, while we estimate $p_i$ directly from the data using the observed features $x_i$ and reoffending outcomes $y_i$. Given a consistent estimator, the probabilities $p_i$ will be perfectly calibrated in the large sample limit, unlike $\hat p_{i,0}$. Our results in Appendix \ref{sec:exp-2-learned} show that the performance of IJDI-Scan is similar with learned and true probabilities $p_i$.

Next we define $FPR(S) = \frac{1}{|S_N|} \sum_{i \in S_N} \hat p_{i,b}$ and $FPR({{\sim}} S) = \frac{1}{|D_N \setminus S_N|} \sum_{i \in D_N \setminus S_N} \hat p_{i,b}$. Using these definitions of the FPR inside and outside $S$, we define the IJDI fairness criterion for FPR as:
\begin{equation}\label{eqn:IJDI-FPR}
FPR(S) - FPR({\sim} S) \le \lambda_N(p(S) - p({\sim} S)),
\end{equation}
if $p(S) > p({\sim} S)$, and $FPR(S) \le FPR({\sim} S)$ otherwise, where $\lambda_N$ is a user-defined constant.

Appendix~\ref{sec:utility} derives the following utility-based formulation of the IJDI fairness criterion:
\begin{align}
\label{eqn:neg-from-utility}
\sum_{i \in S_N} (\hat p_{i,b} - \bar p_b) \le \lambda_N \sum_{i \in S_N} (p_i - \bar p),
\end{align}
where $\lambda_N = \frac{cost(FP)+cost(FN)}{cost(\Delta_{FPR})} 
 = \big(1 + \frac{cost(FN)}{cost(FP)}\big) / {\frac{cost(\Delta_{FPR})}{cost(FP)}}$.
The derivation of this utility-based definition not only justifies the linear form of eqn.~\eqref{eqn:IJDI-FPR}, but also provides an intuitive expression for $\lambda_N$. Please see Appendix~\ref{sec:utility} for further details.  Appendix~\ref{sec:eliciting} provides thorough guidance for choosing a specific value of $\lambda_N$ by eliciting preferences regarding the relative costs of false positives, false negatives, and group-level imbalances in error rates.

The IJDI fairness criterion suggests that differences in FPR up to $\lambda_N$ times the (positive) difference in base rates are considered ``sufficiently justified'', while larger differences are ``insufficiently justified''. Critically, the $\lambda_N$ parameter allows us to smoothly interpolate between the ``error rate balance'' and ``risk-adjusted regression''~\cite{jung2019} fairness criteria.  When $\lambda_N = 0$, the IJDI criterion reduces to error rate balance, identifying any significant differences in FPR as ``unfair''.  When $\lambda_N \rightarrow \infty$, it aligns closely with risk-adjusted regression, since even slight differences in base rates can justify any differences in FPR.   Our results of Experiment 2 below demonstrate that IJDI's detection performance is maximized at intermediate values of $\lambda_N$, thus outperforming both error rate balance and risk-adjusted regression.

\section{IJDI-Scan} \label{sec:ijdi-scan}
We now consider how the above definition of insufficiently justified disparate impact (IJDI) can be used to scan for the intersectional subgroups with the most significant IJDI, thus extending the FPR- and TPR-Scans by adjusting for differences in base rates when scanning for error rate imbalances.  As in the FPR- and TPR-Scans, we can subset the data to the negative individuals $D_N$ or the positive individuals $D_P$, redefine $y_{i,scan}$ and $\hat p_{i,scan}$, pass these values to Bias Scan, and constrain $q>1$ to search for subgroups $S$ with the most significant IJDI.  As before, $y_{i,scan}$ is defined as the binary recommendation $\hat p_{i,b}$, thus representing a false positive or a true positive for individuals $i$ in $D_N$ or $D_P$ respectively.  However, in this section we introduce a new term to the definition of $\hat{p}_{i, scan}$, leading to an extension of the FPR-Scan and TPR-Scan hypothesis tests~(eqn.~\eqref{bias-scan-hypothesis}).

As shown in Appendix~\ref{sec:refactor}, the negative IJDI fairness criterion (eqn.~(\ref{eqn:IJDI-FPR})), 
$FPR(S) - FPR({\sim} S) \le \lambda_N (p(S)-p({\sim} S))$, is equivalent to the criterion derived from our utility-based formulation (eqn.~(\ref{eqn:neg-from-utility})):
$\sum_{i \in S_N} (\hat p_{i,b} - \bar p_b) \le \lambda_N \sum_{i \in S_N} (p_i - \bar p)$, 
where $\bar p_b = \frac{1}{|D_N|} \sum_{i \in D_N} \hat p_{i,b}$ and $\bar p = \frac{1}{|D_N|} \sum_{i \in D_N} p_i$. Rearranging terms, we obtain 
$\sum_{i \in S_N} \hat p_{i,b} \le \sum_{i \in S_N} \left( \bar p_b + \lambda_N(p_i - \bar p) \right)$,
thus allowing us to use $y_{i,scan} = \hat p_{i,b}$ and $\hat p_{i,scan} = \bar p_b + \lambda_N (p_i - \bar p)$ as inputs into Bias Scan, searching for subgroups with significantly higher than expected FPR while adjusting for the base rates (true probabilities) $p_i$.

Substituting the new definition of $\hat{p}_{i, scan}$ into eqn.~\eqref{bias-scan-hypothesis} yields the hypothesis test for IJDI-Scan:
\begin{gather}
    \label{ijdi-scan-hypothesis}
    \begin{split}
        H_0: odds(\hat{p}_{i,b} = 1) &= \frac{\bar{p}_b + \lambda_N (p_i - \bar{p})}{1-(\bar{p}_b + \lambda_N (p_i - \bar{p}))}, \forall i \in D_N, \\
        H_1(S): odds(\hat{p}_{i,b} = 1) &= \frac{q(\bar{p}_b + \lambda_N (p_i - \bar{p}))}{1-(\bar{p}_b + \lambda_N (p_i - \bar{p}))}, \forall i \in S_N, \\
        odds(\hat{p}_{i,b} = 1) &= \frac{\bar{p}_b + \lambda_N (p_i - \bar{p})}{1-(\bar{p}_b + \lambda_N (p_i - \bar{p}))}, \forall i \in D_N \setminus S_N.
    \end{split}
\end{gather}
For the positive IJDI fairness criterion, we obtain the similar expression, $\sum_{i \in S_P} \hat p_{i,b} \le \sum_{i \in S_P} \left( \bar p_b + \lambda_P(p_i - \bar p) \right)$,
where $\bar p_b = \frac{1}{|D_P|} \sum_{i \in D_P} \hat p_{i,b}$ and $\bar p = \frac{1}{|D_P|} \sum_{i \in D_P} p_i$.  This allows us to use $y_{i,scan} = \hat p_{i,b}$ and $\hat p_{i,scan} = \bar p_b + \lambda_P (p_i - \bar p)$ as inputs into the traditional Bias Scan.  It is clear from these definitions that FPR-Scan is a special case of 
IJDI-Scan for negatives with $\lambda_N = 0$, and TPR-Scan is a special case of IJDI-Scan for positives with $\lambda_P = 0$.

While our new IJDI-Scan approach is a convenient extension of the FPR- and TPR-Scans, there are several non-trivial edge cases which must be handled by IJDI-Scan, arising from the newly defined $\hat{p}_{i, scan} = \bar{p}_b + \lambda (p_i - \bar{p})$. These include handling the cases when $p(S) < p({\sim}S)$ and when $\hat{p}_{i,scan}$ falls outside the interval $[0,1]$.  In Appendix~\ref{sec:edge-cases}, we discuss when these edge cases can be violated and how we can make iterative adjustments to address these violations.

In order to properly scan over each subgroup to identify IJDI, we develop an iterative algorithm (Algorithm~\ref{alg:algorithm} in Appendix~\ref{sec:ijdi-algo}) where we first run Bias Scan with the $y_{i, scan}$ and $\hat{p}_{i, scan}$ defined above to detect the intersectional subgroup that most significantly violates the IJDI fairness criterion. If a subgroup is found, we check if either edge case condition is violated, and if so, we make adjustments and re-run Bias Scan. Once no edge case conditions are found to be violated, the scan returns the subgroup with the highest (most significant) log-likelihood ratio score $F(S)$. As in the original Bias Scan, the statistical significance of the detected subgroup can be assessed by randomization testing (described in detail in Appendix~\ref{sec:ijdi-algo}). In the next section, we implement experiments to test the behavior of IJDI-Scan, and we develop strategies for correcting IJDI. 

\section{Experiments} \label{sec:experiments}

Now that we have established the theoretical formulation of the IJDI criterion and developed the IJDI-Scan algorithm described above, we implement two experiments to evaluate how effectively the algorithm detects IJDI. Further, we develop three approaches to mitigating IJDI. 
We conduct experiments using two datasets: ProPublica's COMPAS dataset and the German Credit Dataset. Both datasets are used frequently throughout the fairness literature, and both have categorical features and binary outcomes, making them suitable for scanning over intersectional subgroups to detect error rate imbalances.  These datasets are described in detail in Appendix~\ref{sec:datasets}. 

\subsection{Detecting IJDI}\label{sec:evaluation}
As discussed in Section~\ref{sec:intro}, error rate imbalances can arise from several causes. One factor is the quality of predictions being evaluated: if predictions are systematically biased against a certain subgroup, this will lead to error rate discrepancies between that subgroup and the broader population. However, biased predictions are not the only source of FPR and TPR imbalance~\cite{chouldechova2017fair}. Even if we were to evaluate an oracle prediction -- i.e., the probabilities that truly generated the outcomes -- there may still be significant error rate imbalance if a sharp threshold is used to binarize these predictions. We examine the latter source of IJDI in Experiment 1 and the former in Experiment 2.
\subsubsection{Experiment 1: Detecting IJDI from Sharp Thresholding} In order to examine how accurately IJDI-Scan identifies subgroups that violate the IJDI criterion due to sharp thresholding of predictions, we implement an experiment to perform the scan and compare actual to theoretical results. For this experiment we use the COMPAS dataset features $X_m$, but systematically generate new predictions and prior probabilities $\hat{p}_{i,0}=p_i$. We also generate outcomes $y_{i,0}$ from these predictions to ensure that $\hat{p}_{i,0}=p_i$ are true estimates of $y_{i,0}$. We first draw predictions from uniform distributions, such that $p_i {\sim} [0.51 \pm 0.01k]$ for $i\in S$ and $p_i {\sim} [0.49 \pm 0.01k]$ for $i\not\in S$.  Next, we assume a sharp threshold of $\theta=0.5$ for recommendations: $\hat p_{i,b} = \mathbf{1}\{\hat p_{i,0} > 0.5\}$.  Thus when $k \le 1$, this simplifies to the example given in Section 1 where each individual in $S$ has $\hat p_{i,b}=1$ and all other individuals have $\hat p_{i,b}=0$. Finally, we simulate outcomes $y_{i,0} \:{\sim}\: \text{Bernoulli}(p_i)$.  

When running IJDI-Scan for this experiment, the classifier predictions $\hat{p}_{i,0}$ that the scan audits are equal to the true probabilities $p_i$ that the scan uses to calculate $\hat{p}_{i, scan}$. This ensures that the cause of IJDI is not biased predictions but rather the use of a sharp threshold $\theta$ to binarize those predictions. The benefit of using known distributions to generate predictions and outcomes is that, for a given $k$, we know whether to expect subgroup $S$ to have IJDI for different values of $\lambda$. Namely, we can find the cutoff value $\lambda^\ast$ so that when $\lambda < \lambda^\ast$, we expect $S$ and ${\sim} S$ to have an error rate imbalance that is insufficiently justified by the base rate difference (IJDI), and when $\lambda > \lambda^\ast$, we expect $S$ and ${\sim} S$ to have an error rate imbalance that is sufficiently justified (no IJDI). As shown in Appendix~\ref{sec:theoretical}, the $\lambda^*$ for both the negative and positive IJDI-Scan is
$\lambda^*_N = \lambda^*_P = \frac{75}{k}\frac{4999-k^2}{7497-k^2}$ for $k > 1$, and $\lambda^*_N = \lambda^*_P = 50$ for $0 \le k \le 1$.
%\begin{figure}[t]
%\centering
%\includegraphics[width=0.4\columnwidth]{figures/iou_1_neg.png}
%\includegraphics[width=0.4\columnwidth]{figures/iou_1_pos.png}
%\caption[Experiment 1: IOU as a function of $\lambda$.]{IOU as a function of $\lambda$ for negative IJDI-Scan (left) and positive IJDI-Scan (right). For $k \in \{0,1,3,10\}$, empirical IOU values are plotted and the theoretical cutoff $\lambda^*$ is shown as a vertical dotted line.}
%\label{iou_sim_1}
%\end{figure}
%\begin{figure*}[t]
%\centering
%\includegraphics[width=0.4\columnwidth]{figures/score_1_neg.png}
%\includegraphics[width=0.4\columnwidth]{figures/score_1_pos.png}
%\caption[Experiment 1: Score as a function of $\lambda$.]{Score as a function of $\lambda$ for Experiment 1 negative IJDI-Scan (left) and Experiment 1 positive IJDI-Scan (right).}
%\label{score_sim_1}
%\end{figure*}

\begin{figure}[t]
\centering
(a) \includegraphics[width=0.35\columnwidth]{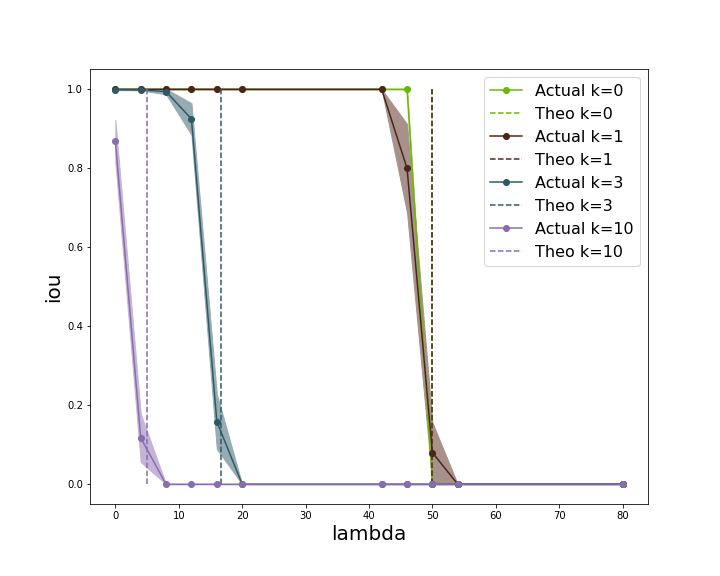}
(b) \includegraphics[width=0.35\columnwidth]{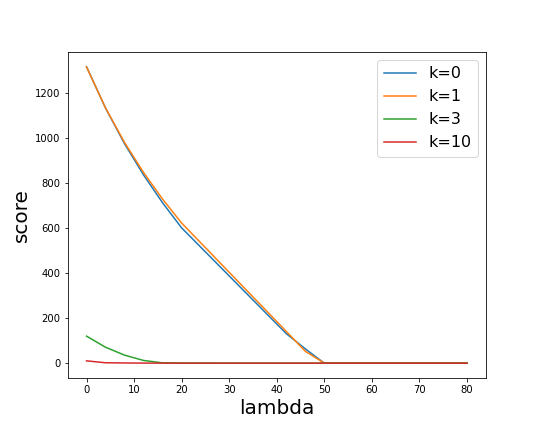}
\caption{Experiment 1 results for negative IJDI-Scan. (a) IOU (with 95\% CI) as a function of $\lambda$. For $k \in \{0,1,3,10\}$, empirical IOU values are plotted and the theoretical cutoff $\lambda^*$ is shown as a vertical dotted line. (b) Score as a function of $\lambda$, for $k \in \{0,1,3,10\}$.}
\label{fig:sim_1_neg}
\end{figure}

For different values of $k$ and $\lambda$, we run IJDI-Scan 100 times each, computing the intersection-over-union ($IOU$) of $S^*$ with $S$, $IOU = \frac{|S^* \cap S|}{|S^* \cup S|}$, each time. We report the average $IOU$ as well as the average score $F(S^*)$ over these iterations. Figure~\ref{fig:sim_1_neg} shows the results for the negative IJDI-Scan, and results for the positive IJDI-Scan are provided in Appendix~\ref{sec:pos-experiments}, Figure~\ref{fig:sim_1_pos}.  As Figures~\ref{fig:sim_1_neg}(a) and~\ref{fig:sim_1_pos}(a) demonstrate for various values of $k$, our empirical results align well with the theoretical value of $\lambda^*$: for $\lambda < \lambda^*$, the subgroup found by IJDI-Scan has high overlap with $S$, while for $\lambda \ge \lambda^*$, the overlap is close to 0.  Similarly, as shown in Figures~\ref{fig:sim_1_neg}(b) and~\ref{fig:sim_1_pos}(b), the score of the detected subgroup is decreasing with $\lambda$, and close to 0 for $\lambda \ge \lambda^*$.  These results demonstrate the ability of IJDI-Scan to accurately detect violations of the IJDI criteria, and the use of the $\lambda$ parameter to specify how much imbalance in error rates is justified by a given difference in base rates.

%Figure~\ref{score_sim_1} shows the IJDI-Scan log-likelihood ratio scores of the detected subsets, $F(S^\ast)$, as a function of $\lambda$. The score is for the most part monotonically decreasing in $\lambda$ and $k$ for both the negative and positive scan, as expected. The one exception is that the score curves for $k=1$ are slightly higher than the score curves for $k=0$, which is as expected from our theoretical results above.  

%For the negative IJDI-Scan, both $k=0$ and $k=1$ have $FPR(S)-FPR({\sim} S) = 1$, but $p(S)-p({\sim} S)$ is slightly lower for $k=1$ because of the preferential selection of individuals who have lower probabilities $p_i$ into the negative class. As shown in our theoretical derivation, both $p(S)$ and $p({\sim} S)$ are reduced as compared to the $k=0$ case, but $p(S)$ is reduced slightly more than $p({\sim} S)$, leading to $p(S)-p({\sim} S)$ slightly less than 0.02.  

%Similarly, for the positive IJDI-Scan, both $k=0$ and $k=1$ have $TPR(S)-TPR({\sim} S) = 1$, but again $p(S)-p({\sim} S)$ is slightly lower for $k=1$, this time because of the preferential selection of individuals who have higher probabilities $p_i$ into the positive class.  Both $p(S)$ and $p({\sim} S)$ are increased as compared to the $k=0$ case, but $p({\sim} S)$ is increased slightly more than $p(S)$, again leading to $p(S)-p({\sim} S)$ slightly less than 0.02.

\subsubsection{Experiment 2: Detecting IJDI from Biased Predictions} Next we explore how well IJDI-Scan can identify subgroups with systematically biased predictions, using realistic probabilities of the outcome $y_{i,0}$ (reoffending) based on the actual COMPAS data. We build a logistic regression model trained on the entire COMPAS dataset for $X_m \setminus X_s$, where values for the features $X_s$ (age, sex, and race) are randomly chosen and uniquely identify subgroup $S$. We generate predictions for each individual to use as the prior probabilities $p_i$, and as in Experiment 1, we generate outcomes $y_{i,0} \:{\sim}\: \text{Bernoulli}(p_i)$. We set $logit(\hat{p}_{i,0}):=logit(p_i) + \gamma$ for $i\in S$, where $\gamma$ represents a log-odds shift, and $\hat{p}_{i,0}:=p_i$ for $i \not\in S$. As we show in Appendix~\ref{sec:shift}, the log-odds shift for $i \in S$ is equivalent to setting $\hat{p}_{i,0} := \frac{p_i e^{\gamma}}{p_i e^{\gamma} + 1 - p_i}.$ Since we are systematically skewing $\hat{p}_{i,0}$ upward for $S$, which should increase the instances of $\hat{p}_{i,b}=1$, $S$ should be more likely to be returned by both the negative and positive IJDI-Scan. Specifically, for larger $\gamma$ we expect that a larger $\lambda$ would be required to sufficiently justify the error rate imbalance resulting from biased predictions.

The impact of a log-odds shift upward for $\hat{p}_{i,0}$ should be equivalent to that of a log-odds shift downward for the threshold $\theta_i$, as both shifts would increase the instances of $\hat{p}_{i,b}=1$ for $S$ by the same quantity. Thus, instead of shifting $logit(\hat{p}_{i,0})$ upward by $\gamma$,
we can shift $logit(\theta_i)$ downward by the same amount. Assuming a constant initial threshold of $\theta_0$, we can therefore set 
$\theta_i := \frac{\theta_0 e^{-\gamma}}{\theta_0 e^{-\gamma} + 1 - \theta_0}, \forall i \in S,$
and set $\theta_i:=\theta_0, \forall i \not\in S$. We examine the results for both types of shift.

%\begin{figure}[t]
%\centering
%\includegraphics[width=0.4\columnwidth]{figures/iou_2_neg.png}
%\includegraphics[width=0.4\columnwidth]{figures/iou_2_pos.png}
%\caption{IOU as a function of $\lambda$ for negative IJDI-Scan (left) and positive IJDI-Scan (right). For $\gamma \in \{0,0.3, 3\}$, IOU values are plotted for an upward log-odds shift of probability (triangle) and downward log-odds shift of threshold (x).}
%\label{sim_2}
%\end{figure}
%\begin{figure*}[t]
%\centering
%\includegraphics[width=0.4\columnwidth]{figures/score_2_neg.png}
%\includegraphics[width=0.4\columnwidth]{figures/score_2_pos.png}
%\caption[Experiment 2: Score as a function of $\lambda$.]{Score as a function of $\lambda$ for Experiment 2 negative IJDI-Scan (left) and Experiment 2 positive IJDI-Scan (right).}
%\label{score_sim_2}
%\end{figure*}

\begin{figure}[t]
\centering
(a) \includegraphics[width=0.35\columnwidth]{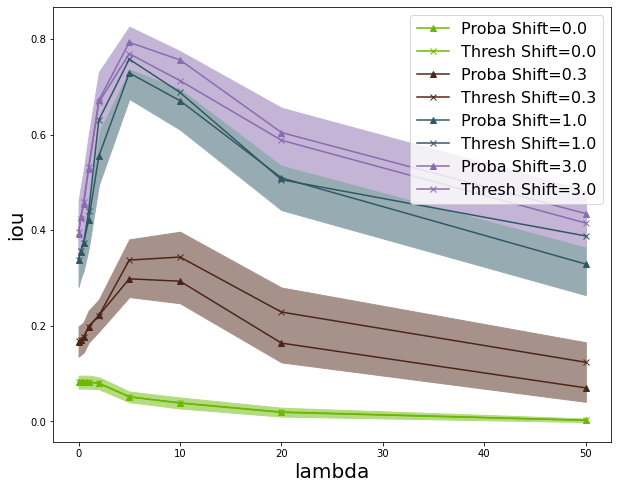}
(b) \includegraphics[width=0.35\columnwidth]{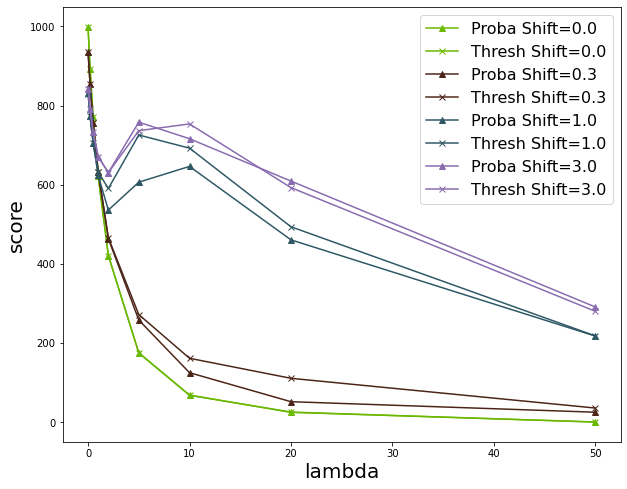}
\caption{Experiment 2 results for negative IJDI-Scan. (a) IOU (with 95\% CI) as a function of $\lambda$. For $\gamma \in \{0,0.3, 1, 3\}$, IOU values are plotted for an upward log-odds shift of probability (triangle) and downward log-odds shift of threshold (x). (b) Score as a function of $\lambda$, for $\gamma \in \{0,0.3, 1, 3\}$.}
\label{fig:sim_2_neg}
\end{figure}

We run IJDI-Scan 200 times for each $\gamma$ and $\lambda$. We show results for negative IJDI-Scan in Figure~\ref{fig:sim_2_neg} and for positive IJDI-Scan in Appendix~\ref{sec:pos-experiments}, Figure~\ref{fig:sim_2_pos}.  In Figures~\ref{fig:sim_2_neg}(a) and~\ref{fig:sim_2_pos}(a), we see that larger predictive biases $\gamma$ yield a larger $IOU$ between $S^*$ and $S$, as expected. The relationship between $\lambda$ and $IOU$ is more interesting: for subtle signals, IJDI-Scan with an intermediate value of $\lambda$ is able to detect the bias more accurately than either FPR-Scan / TPR-Scan ($\lambda=0$) or risk-adjusted regression (large $\lambda$). This is because the COMPAS dataset has demographics with widely varying base rates, and resulting disparities in both FPR and TPR.  If these base rate differences are not sufficiently accounted for (i.e., if $\lambda$ is too small), these subgroups score higher than $S$, while for moderate $\lambda$, these subgroups' disparate impact is justified by the base rate differences, reducing their scores and allowing $S$ to be detected. For large $\lambda$, subgroup $S$ is not detected by IJDI-Scan as there is little or no IJDI present. As Figures~\ref{fig:sim_2_neg}(b) and~\ref{fig:sim_2_pos}(b) show, the score $F(S^\ast)$ generally decreases with $\lambda$ and increases with signal strength $\gamma$.  Finally, as a robustness check, we re-run Experiment 2 using probabilities $p_i$ learned from the data rather than the true probabilities $p_i$, and observe nearly identical results (see Appendix~\ref{sec:exp-2-learned}).

\subsection{Mitigating IJDI, Approach 1: Correction for a Specific Subgroup}\label{sec:mit-1} 
\begin{figure}[t]
\centering
(a) \includegraphics[width=0.35\columnwidth]{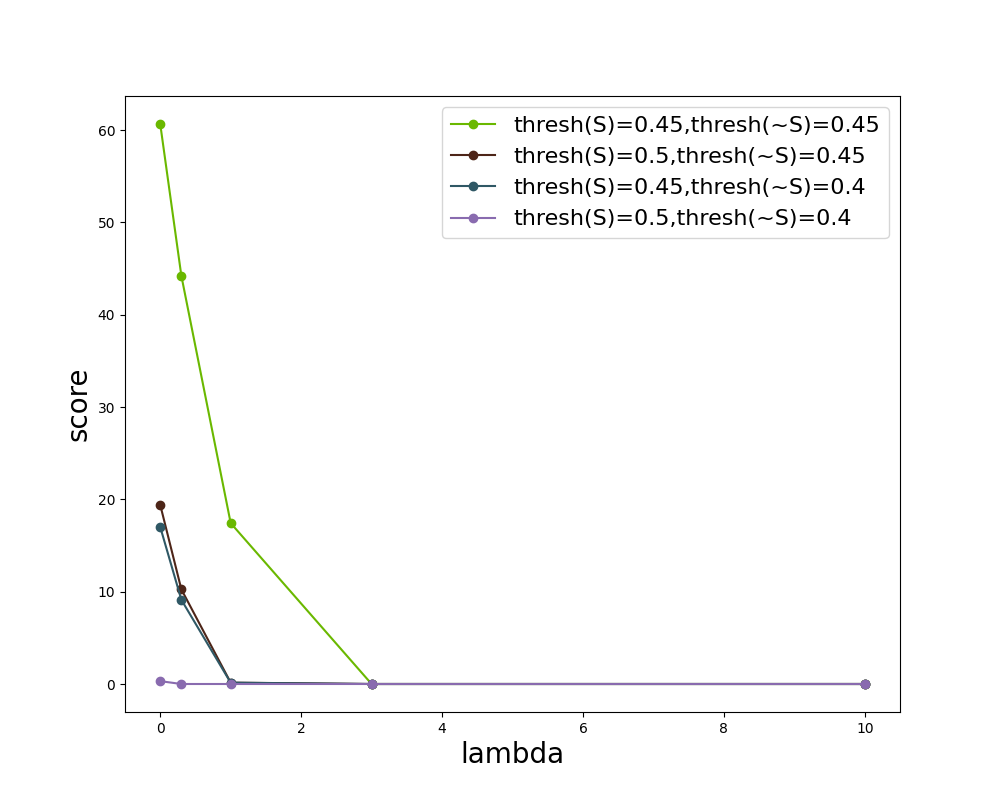}
(b) \includegraphics[width=0.35\columnwidth]{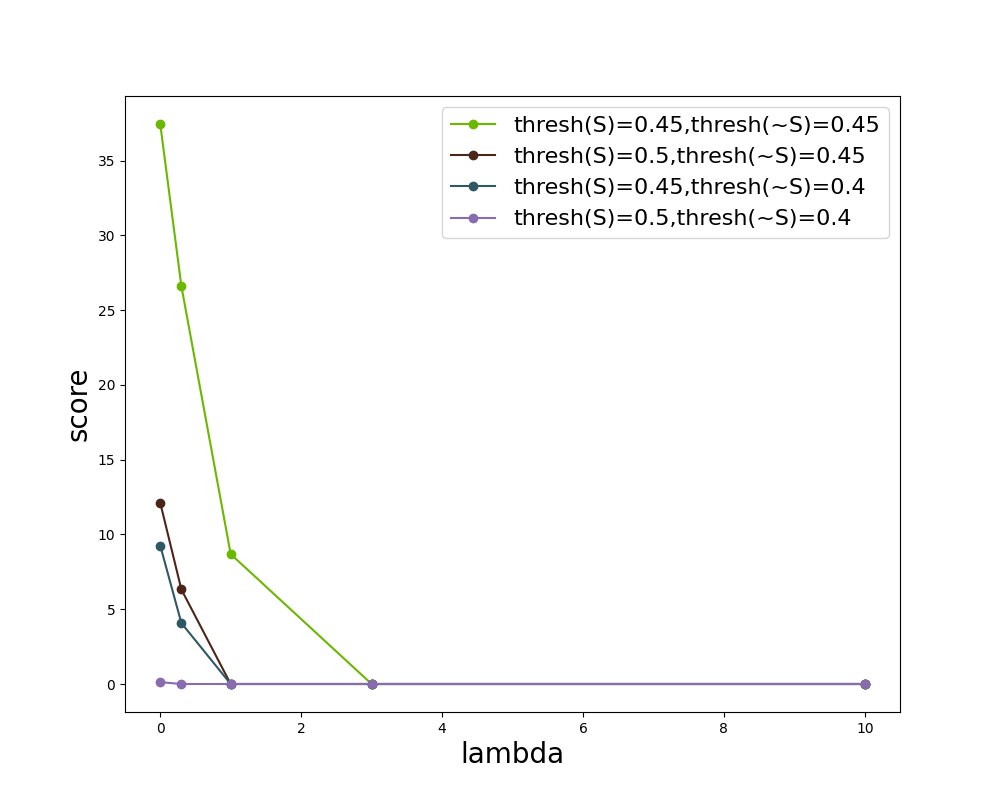}
\caption{Mitigation approach 1 results. Score as a function of $\lambda$ for (a) negative IJDI-Scan and (b) positive IJDI-Scan. Score curves for each case described in Table~\ref{table:compas-error-table} are plotted.}
\label{fig:score_mit_1}
\end{figure}

\citet{chouldechova2017fair} demonstrated that, on the COMPAS dataset, Black defendants had higher FPR and TPR even when predictions were correctly calibrated, and noted that using different thresholds $\theta$ for Black and non-Black defendants could potentially mitigate this disparity.  We use the COMPAS data to test four cases of differing thresholds between $S=$\{``race": [``African-American"]\} and ${\sim} S$. A breakdown of these cases as well as FPR and TPR comparison across groups is shown in Table~\ref{table:compas-error-table}. We can see the largest error rate discrepancy when the threshold is constant across groups. As we expect, this discrepancy is drastically reduced by group-dependent thresholds. Figure~\ref{fig:score_mit_1} demonstrates that a larger $\lambda$ is required to justify a larger error rate imbalance: 
for a constant threshold, the IJDI criteria are violated for $\lambda < 3$, while group-dependent thresholds give low IJDI-Scan scores even for $\lambda \approx 0$.  
\begin{table}
  \begin{center}
    \begin{tabular}{cccccc} 
     \textbf{$\theta_i,\forall i \in S$} & \textbf{$\theta_i,\forall i \in {\sim} S$} & \textbf{$FPR(S)$} & \textbf{$FPR({\sim} S)$} & \textbf{$TPR(S)$} & \textbf{$TPR({\sim} S)$}\\
     \hline
     0.45 & 0.45 & 44.8\% & 22.0\% & 72.0\% & 49.3\% \\ 
     0.5 & 0.45 & 34.3\% & 22.0\% & 62.8\% & 49.3\% \\ 
     0.45 & 0.4 & 44.8\% & 32.5\% & 72.0\% & 61.0\% \\ 
     0.5 & 0.4 & 34.3\% & 32.5\% & 62.8\% & 61.0\% \\ 
    \end{tabular}
  \end{center}
  \caption{Risk thresholds and resulting error rates for mitigation approach 1.  Experiments on COMPAS data, using different risk thresholds for Black and non-Black defendants.}
  \label{table:compas-error-table}
  \vspace{-3mm}
\end{table}

\subsection{Mitigating IJDI, Approach 2: Iterative Correction}\label{sec:mit-2}  We propose another approach in which we adjust $\theta_i$ for individuals in subgroup $S^*$ returned by IJDI-Scan and iteratively re-run the scan until no subgroup is returned. The null hypothesis assumes $PR(S^*) \le PR({\sim} S^*) + \lambda (p(S^*) - p({\sim} S^*)),$ where $PR$ is $FPR$ for negative IJDI-Scan and $TPR$ for positive IJDI-Scan. Thus, to ensure that $S^*$ is no longer returned by the scan, we can set $\theta_i := \eta(S^*), \forall i \in S^*,$ where $\eta(S)$ is the $[1 - (PR({\sim} S) + \lambda (p(S) - p({\sim} S))]$ quantile of $\hat{p}_{i, 0}$, for $i \in S$.  This choice of threshold guarantees that no more than $PR({\sim} S^\ast) + \lambda (p(S^\ast) - p({\sim} S^\ast))$ proportion of the $\hat p_{i,0}$ values in $S^\ast$ exceed $\theta_i$, and thus, that no IJDI is present.

%\begin{figure}[t]
%\centering
%\includegraphics[width=0.4\columnwidth]{figures/score_compas_mit_2_neg.png}
%\includegraphics[width=0.4\columnwidth]{figures/score_compas_mit_2_pos.png} \\
%\caption[Mitigation Approach 2: Score as a function of number of corrections for COMPAS.]{Score on COMPAS as a function of the number of corrections for negative IJDI-Scan (left) and positive IJDI-Scan (right).}
%\label{score_mit_2_compas}
%\end{figure}
%\begin{figure}[t]
%\centering
%\includegraphics[width=0.4\columnwidth]{figures/score_german_credit_mit_2_neg.png}
%\includegraphics[width=0.4\columnwidth]{figures/score_german_credit_mit_2_pos.png}
%\caption[Mitigation Approach 2: Score as a function of number of corrections for German Credit.]{Score on German credit as a function of the number of corrections for negative IJDI-Scan (left) and positive IJDI-Scan (right).}
%\label{score_mit_2_german_credit}
%\end{figure}

\begin{figure}[t]
\centering
(a) \includegraphics[width=0.35\columnwidth]{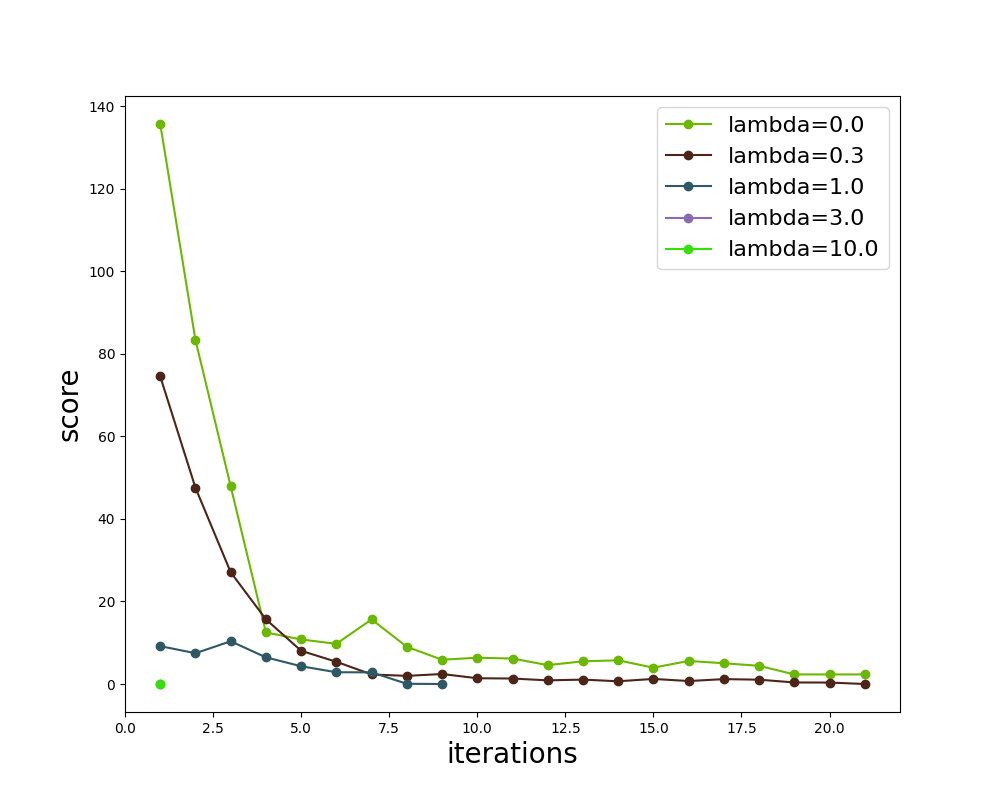}
(b) \includegraphics[width=0.35\columnwidth]{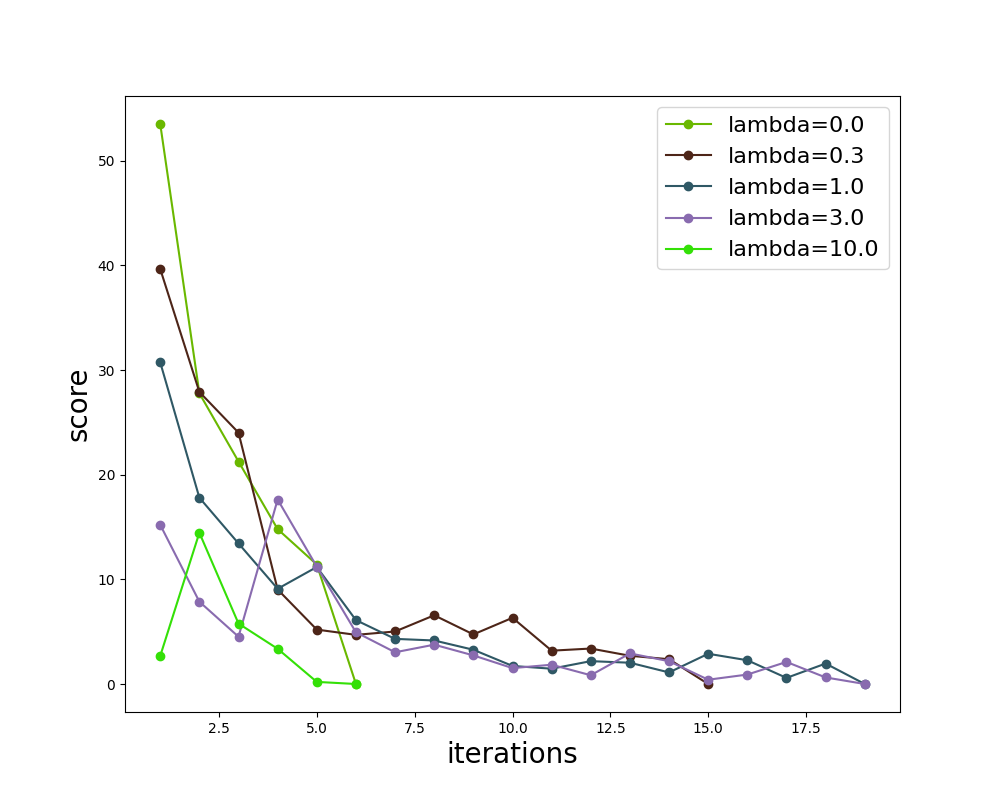} \\
\caption{Mitigation approach 2 results for negative IJDI-Scan. Score on (a) COMPAS and (b) German credit, as a function of the number of corrections.}
\label{fig:score_mit_2_neg}
\end{figure}

We implement this technique on both the COMPAS and German Credit datasets. For the former, we use the COMPAS risk scores as predictions $\hat{p}_{i,0}$ and the logistic regression model from Experiment 2 as prior probabilities $p_i$. For the latter, we train logistic regression and random forest models on the dataset; we use the former as $\hat{p}_{i,0}$ and the latter as $p_i$. Results for negative IJDI-Scan are shown in Figure~\ref{fig:score_mit_2_neg},  and results for positive IJDI-Scan are shown in Appendix~\ref{sec:pos-experiments}, Figure~\ref{fig:score_mit_2_pos}. 
We observe that the score of the subset detected by IJDI-Scan generally decreases as more subgroups are corrected. However, it is possible for correction of one subgroup to increase IJDI for another subgroup, by reducing the FPR or TPR outside the latter subgroup. For each dataset, we report the first subgroups that were detected and mitigated in Appendix~\ref{sec:report-subgroups}, for both negative and positive IJDI-Scans. While IJDI-Scans using various $\lambda$ values found similar disparate impacts affecting subgroups with higher base rates (e.g., against younger and Black defendants for COMPAS), allowing multiple corrections and a larger $\lambda$ also revealed more subtle, intersectional biases (e.g., against younger females).

\subsection{Mitigating IJDI, Approach 3: Randomization of Thresholds}
\label{sec:mit-3}
One possible approach to mitigating IJDI resulting from sharp thresholds is randomization of the threshold $\theta$, which is used to map the classifier's probabilistic predictions $\hat p_{i,0}$ to binary recommendations $\hat p_{i,b}$. For example, in the example in Section~\ref{sec:intro} where $p_i=0.51, \forall i \in S$ and $p_i=0.49, \forall i \not\in S$, instead of using a constant threshold $\theta=0.5$ we could draw thresholds $\theta_i$ uniformly on $[\theta-\Delta, \theta+\Delta]$, $\Delta \le 0.5$. As we demonstrate in Appendix~\ref{sec:randomization}, this randomization approach guarantees that no IJDI is present for any $\lambda \ge \frac{1}{2\Delta}$. Randomization can reduce subgroup-level biases but increases variance between individual outcomes, making it undesirable to inject. But it is worth noting that some randomization already exists in practice, e.g., defendants are randomly assigned to harsher or more lenient judges, and this randomness may be beneficial for mitigating IJDI.

\section{Related Work}

As discussed above, our IJDI-Scan approach, as well as our simpler FPR- and TPR-Scans (which do not take base rate differences into account), extend the previous Bias Scan approach~\cite{biasscan}, which was originally proposed only for calibration, to new definitions of fairness including the novel IJDI criterion proposed here.  Bias Scan is one of several recently proposed approaches for auditing subgroup fairness~\cite{biasscan,kearns2018preventing,hebert-johnson18a,kim2019multiaccuracy}.
IJDI interpolates between strict error rate balance, i.e., equalized odds and equality of opportunity~\cite{hardt2016}, and the criterion of Jung et al.~\cite{jung2019}, which justifies \emph{any} error rate difference if a corresponding base rate difference is present. The IJDI criterion was motivated by the issues with strict error rate balance raised by~\cite{corbett2018measure} and~\cite{hu2020}, as described in Section~\ref{sec:intro}.

Several previous papers~\cite{bilalzafar2019,pleiss2017,hu2020} have considered adding ``slack'' to standard fairness definitions, allowing criteria such as error rate balance to be approximately rather than exactly met.  Bilal-Zafar et al.~\cite{bilalzafar2019} consider learning margin-based classifiers while enforcing approximate statistical parity constraints like the ``80\% rule'', and evaluate tradeoffs between fairness and accuracy.  Similarly, Hu \& Chen~\cite{hu2020} consider $\epsilon$-fairness (an additive approximation of balanced error rates) and show that overly tight restrictions on $\epsilon$ lead to solutions that are Pareto-suboptimal with respect to welfare maximization.  Unlike IJDI, these previous definitions cannot easily be applied to audit classifiers for subgroup fairness.  A single constraint on how much difference in FPR or TPR is ``allowable'' does not allow comparison of subgroups of different sizes: for example, a 100\% difference in FPR between subgroups would be much less meaningful if each subgroup contained only a single individual. Moreover, the previous definitions do not account for base rate differences between subgroups.  For example, a 20\% difference in FPR might be acceptable for two subgroups with very different base rates (99\% vs. 1\%) and unacceptable for two subgroups with similar base rates (51\% vs. 49\%).

Finally, Pleiss et al.~\cite{pleiss2017} consider a ``balanced cost'' criterion which trades off between false positive and false negative errors.  While this bears some resemblance to our utility-based derivation of IJDI, the two notions of fairness are entirely different.
First, Pleiss et al. use a definition of “generalized” FPR and FNR which does not account for the binarization of a classifier’s probabilistic predictions. Thus they cannot identify unfairness resulting from sharp thresholding of predictions, a primary motivation for IJDI.
Second, they rely heavily on the assumption that the classifier is perfectly calibrated, and thus has a simple relationship between its generalized FPR and FNR.  Thus they consider an idealized, theoretical scenario rather than providing a practical tool for auditing classifiers as we do.
Finally, ``balanced cost'' assumes that FPR and FNR have the same type of impact on the affected group (with possibly different weights), allowing FPR imbalances and FNR imbalances to trade off. In other words, Pleiss et al. might consider a scenario where one group gets all the false positives and the other group gets all the false negatives to be fair, while IJDI would not.

\section{Conclusions}

In this paper, we derived a new fairness criterion that allows us to measure error rate imbalances while also incorporating base rates. The new criterion includes a parameter $\lambda$ that allows a policy-maker to determine the degree to which differences in base rates between groups can justify error rate disparities. In order to efficiently identify intersectional subgroups which violate our IJDI criterion, we developed a novel technique, IJDI-Scan, which builds on two new extensions of Bias Scan: FPR- and TPR-Scan. Our experiments show that IJDI-Scan effectively detects IJDI originating from both sharp thresholds and biased predictions. We also explored three ways of mitigating IJDI -- directly modifying thresholds, iteratively adjusting thresholds for detected subgroups, and randomizing thresholds.  Limitations of the work are discussed in Appendix~\ref{sec:limitations}.  Nevertheless, we believe that IJDI-Scan provides a useful tool for identifying error rate imbalances while accounting for base rates.

\begin{ack}
This material is based upon work supported by the National Science Foundation Program on Fairness in Artificial Intelligence in Collaboration with Amazon, grant IIS-2040898. Any opinions, findings, and conclusions or recommendations expressed in this material are those of the authors and do not necessarily reflect the views of the National Science Foundation or Amazon.
\end{ack}

% \section{Supplementary Material}

% Authors may wish to optionally include extra information (complete proofs, additional experiments and plots) in the appendix. All such materials should be part of the supplemental material (submitted separately) and should NOT be included in the main submission.

\bibliographystyle{ACM-Reference-Format}
\bibliography{neurips_2023}


\begin{thebibliography}{17}

%%% the Web address.

\ifx \showCODEN    \undefined \def \showCODEN     #1{\unskip}     \fi
\ifx \showDOI      \undefined \def \showDOI       #1{#1}\fi
\ifx \showISBNx    \undefined \def \showISBNx     #1{\unskip}     \fi
\ifx \showISBNxiii \undefined \def \showISBNxiii  #1{\unskip}     \fi
\ifx \showISSN     \undefined \def \showISSN      #1{\unskip}     \fi
\ifx \showLCCN     \undefined \def \showLCCN      #1{\unskip}     \fi
\ifx \shownote     \undefined \def \shownote      #1{#1}          \fi
\ifx \showarticletitle \undefined \def \showarticletitle #1{#1}   \fi
\ifx \showURL      \undefined \def \showURL       {\relax}        \fi
% invisible to TeX
\providecommand\bibfield[2]{#2}
\providecommand\bibinfo[2]{#2}
\providecommand\natexlab[1]{#1}
\providecommand\showeprint[2][]{arXiv:#2}

\bibitem[Angwin et~al\mbox{.}(2016)]%
        {angwin2016machine}
\bibfield{author}{\bibinfo{person}{Julia Angwin}, \bibinfo{person}{Jeff
  Larson}, \bibinfo{person}{Surya Mattu}, {and} \bibinfo{person}{Lauren
  Kirchner}.} \bibinfo{year}{2016}\natexlab{}.
\newblock \showarticletitle{Machine bias}.
\newblock In \bibinfo{booktitle}{\emph{Ethics of Data and Analytics}}.
  \bibinfo{publisher}{Auerbach Publications}, \bibinfo{pages}{254--264}.
\newblock


\bibitem[Bolukbasi et~al\mbox{.}(2016)]%
        {bolukbasi16}
\bibfield{author}{\bibinfo{person}{T. Bolukbasi}, \bibinfo{person}{K.~W.
  Chang}, \bibinfo{person}{J.~Y. Zou}, \bibinfo{person}{V. Saligrama}, {and}
  \bibinfo{person}{A.~T. Kalai}.} \bibinfo{year}{2016}\natexlab{}.
\newblock \showarticletitle{Man is to computer programmer as woman is to
  homemaker? Debiasing word embeddings}.
\newblock In \bibinfo{booktitle}{\emph{Advances in Neural Information
  Processing Systems}}. \bibinfo{pages}{4349--4357}.
\newblock


\bibitem[Buolamwini and Gebru(2018)]%
        {buolamwini18}
\bibfield{author}{\bibinfo{person}{J. Buolamwini} {and} \bibinfo{person}{T.
  Gebru}.} \bibinfo{year}{2018}\natexlab{}.
\newblock \showarticletitle{Gender shades: intersectional accuracy disparities
  in commercial gender classification}.
\newblock In \bibinfo{booktitle}{\emph{Proc. 1st Conf. on Fairness,
  Accountability and Transparency, PMLR 81}}. \bibinfo{pages}{77--91}.
\newblock


\bibitem[Chouldechova(2017)]%
        {chouldechova2017fair}
\bibfield{author}{\bibinfo{person}{Alexandra Chouldechova}.}
  \bibinfo{year}{2017}\natexlab{}.
\newblock \showarticletitle{Fair prediction with disparate impact: A study of
  bias in recidivism prediction instruments}.
\newblock \bibinfo{journal}{\emph{Big Data}} \bibinfo{volume}{5},
  \bibinfo{number}{2} (\bibinfo{year}{2017}), \bibinfo{pages}{153--163}.
\newblock


\bibitem[Corbett-Davies and Goel(2018)]%
        {corbett2018measure}
\bibfield{author}{\bibinfo{person}{Sam Corbett-Davies} {and}
  \bibinfo{person}{Sharad Goel}.} \bibinfo{year}{2018}\natexlab{}.
\newblock \showarticletitle{The measure and mismeasure of fairness: A critical
  review of fair machine learning}.
\newblock \bibinfo{journal}{\emph{arXiv preprint arXiv:1808.00023}}
  (\bibinfo{year}{2018}).
\newblock


\bibitem[Corbett-Davies et~al\mbox{.}(2017)]%
        {corbett-davies17}
\bibfield{author}{\bibinfo{person}{Sam Corbett-Davies}, \bibinfo{person}{Emma
  Pierson}, \bibinfo{person}{Avi Feller}, \bibinfo{person}{Sharad Goel}, {and}
  \bibinfo{person}{Aziz Huq}.} \bibinfo{year}{2017}\natexlab{}.
\newblock \showarticletitle{Algorithmic Decision Making and the Cost of
  Fairness}. In \bibinfo{booktitle}{\emph{Proc. 23rd Conf. on Knowledge
  Discovery and Data Mining}}.
\newblock


\bibitem[Gelman et~al\mbox{.}(2007)]%
        {gelman07}
\bibfield{author}{\bibinfo{person}{A. Gelman}, \bibinfo{person}{J. Fagan},
  {and} \bibinfo{person}{A. Kiss}.} \bibinfo{year}{2007}\natexlab{}.
\newblock \showarticletitle{An analysis of the {New York City Police
  Department's} ``stop-and-frisk'' policy in the context of claims of racial
  bias}.
\newblock \bibinfo{journal}{\emph{J. Amer. Statist. Assoc.}}
  \bibinfo{volume}{102} (\bibinfo{year}{2007}), \bibinfo{pages}{813--823}.
\newblock


\bibitem[Hardt et~al\mbox{.}(2016)]%
        {hardt2016}
\bibfield{author}{\bibinfo{person}{M. Hardt}, \bibinfo{person}{E. Price}, {and}
  \bibinfo{person}{N. Srebro}.} \bibinfo{year}{2016}\natexlab{}.
\newblock \showarticletitle{Equality of opportunity in supervised learning}.
\newblock In \bibinfo{booktitle}{\emph{Advances in Neural Information
  Processing Systems}}. \bibinfo{pages}{3323--3331}.
\newblock


\bibitem[Hebert-Johnson et~al\mbox{.}(2018)]%
        {hebert-johnson18a}
\bibfield{author}{\bibinfo{person}{Ursula Hebert-Johnson},
  \bibinfo{person}{Michael Kim}, \bibinfo{person}{Omer Reingold}, {and}
  \bibinfo{person}{Guy Rothblum}.} \bibinfo{year}{2018}\natexlab{}.
\newblock \showarticletitle{Multicalibration: Calibration for the
  ({C}omputationally-Identifiable) Masses}. In
  \bibinfo{booktitle}{\emph{Proceedings of the 35th International Conference on
  Machine Learning}} \emph{(\bibinfo{series}{Proceedings of Machine Learning
  Research}, Vol.~\bibinfo{volume}{80})},
  \bibfield{editor}{\bibinfo{person}{Jennifer Dy} {and}
  \bibinfo{person}{Andreas Krause}} (Eds.). \bibinfo{publisher}{PMLR},
  \bibinfo{pages}{1939--1948}.
\newblock


\bibitem[Hu and Chen(2020)]%
        {hu2020}
\bibfield{author}{\bibinfo{person}{Lily Hu} {and} \bibinfo{person}{Yiling
  Chen}.} \bibinfo{year}{2020}\natexlab{}.
\newblock \showarticletitle{Fair Classification and Social Welfare}.
\newblock In \bibinfo{booktitle}{\emph{Proc. Conference on Fairness,
  Accountability, and Transparency}}. \bibinfo{pages}{535--545}.
\newblock


\bibitem[Jung et~al\mbox{.}(2019)]%
        {jung2019}
\bibfield{author}{\bibinfo{person}{J. Jung}, \bibinfo{person}{S.
  Corbett-Davies}, \bibinfo{person}{R. Shroff}, {and} \bibinfo{person}{S.
  Goel}.} \bibinfo{year}{2019}\natexlab{}.
\newblock \bibinfo{title}{Omitted and included variable bias in tests for
  disparate impact. Working paper}.
\newblock
\newblock


\bibitem[Kearns et~al\mbox{.}(2018)]%
        {kearns2018preventing}
\bibfield{author}{\bibinfo{person}{Michael Kearns}, \bibinfo{person}{Seth
  Neel}, \bibinfo{person}{Aaron Roth}, {and} \bibinfo{person}{Zhiwei~Steven
  Wu}.} \bibinfo{year}{2018}\natexlab{}.
\newblock \showarticletitle{Preventing fairness gerrymandering: Auditing and
  learning for subgroup fairness}. In \bibinfo{booktitle}{\emph{International
  Conference on Machine Learning}}. PMLR, \bibinfo{pages}{2564--2572}.
\newblock


\bibitem[Kim et~al\mbox{.}(2019)]%
        {kim2019multiaccuracy}
\bibfield{author}{\bibinfo{person}{Michael~P Kim}, \bibinfo{person}{Amirata
  Ghorbani}, {and} \bibinfo{person}{James Zou}.}
  \bibinfo{year}{2019}\natexlab{}.
\newblock \showarticletitle{Multiaccuracy: Black-box post-processing for
  fairness in classification}. In \bibinfo{booktitle}{\emph{Proceedings of the
  2019 AAAI/ACM Conference on AI, Ethics, and Society}}. ACM,
  \bibinfo{pages}{247--254}.
\newblock


\bibitem[Mehrabi et~al\mbox{.}(2021)]%
        {mehrabi2021survey}
\bibfield{author}{\bibinfo{person}{Ninareh Mehrabi}, \bibinfo{person}{Fred
  Morstatter}, \bibinfo{person}{Nripsuta Saxena}, \bibinfo{person}{Kristina
  Lerman}, {and} \bibinfo{person}{Aram Galstyan}.}
  \bibinfo{year}{2021}\natexlab{}.
\newblock \showarticletitle{A survey on bias and fairness in machine learning}.
\newblock \bibinfo{journal}{\emph{Comput. Surveys}} \bibinfo{volume}{54},
  \bibinfo{number}{6} (\bibinfo{year}{2021}), \bibinfo{pages}{1--35}.
\newblock


\bibitem[Pleiss et~al\mbox{.}(2017)]%
        {pleiss2017}
\bibfield{author}{\bibinfo{person}{Geoff Pleiss}, \bibinfo{person}{Manish
  Raghavan}, \bibinfo{person}{Felix Wu}, \bibinfo{person}{Jon Kleinberg}, {and}
  \bibinfo{person}{Kilian~Q Weinberger}.} \bibinfo{year}{2017}\natexlab{}.
\newblock \showarticletitle{On Fairness and Calibration}. In
  \bibinfo{booktitle}{\emph{Advances in Neural Information Processing
  Systems}}, \bibfield{editor}{\bibinfo{person}{I.~Guyon},
  \bibinfo{person}{U.~Von Luxburg}, \bibinfo{person}{S.~Bengio},
  \bibinfo{person}{H.~Wallach}, \bibinfo{person}{R.~Fergus},
  \bibinfo{person}{S.~Vishwanathan}, {and} \bibinfo{person}{R.~Garnett}}
  (Eds.), Vol.~\bibinfo{volume}{30}. \bibinfo{publisher}{Curran Associates,
  Inc.}
\newblock
\urldef\tempurl%
\url{https://proceedings.neurips.cc/paper/2017/file/b8b9c74ac526fffbeb2d39ab038d1cd7-Paper.pdf}
\showURL{%
\tempurl}


\bibitem[Zafar et~al\mbox{.}(2019)]%
        {bilalzafar2019}
\bibfield{author}{\bibinfo{person}{Muhammad~Bilal Zafar},
  \bibinfo{person}{Isabel Valera}, \bibinfo{person}{Manuel Gomez-Rodriguez},
  {and} \bibinfo{person}{Krishna~P. Gummadi}.} \bibinfo{year}{2019}\natexlab{}.
\newblock \showarticletitle{Fairness Constraints: A Flexible Approach for Fair
  Classification}.
\newblock \bibinfo{journal}{\emph{Journal of Machine Learning Research}}
  \bibinfo{volume}{20}, \bibinfo{number}{75} (\bibinfo{year}{2019}),
  \bibinfo{pages}{1--42}.
\newblock


\bibitem[Zhang and Neill(2016)]%
        {biasscan}
\bibfield{author}{\bibinfo{person}{Zhe Zhang} {and} \bibinfo{person}{Daniel~B.
  Neill}.} \bibinfo{year}{2016}\natexlab{}.
\newblock \showarticletitle{Identifying significant predictive bias in
  classifiers}.
\newblock \bibinfo{journal}{\emph{Presented at Workshop on Fairness,
  Accountability, and Transparency in Machine Learning (FAT/ML). arXiv preprint
  arXiv:1611.08292}} (\bibinfo{year}{2016}).
\newblock


\end{thebibliography}

\newpage

\appendix

\section{IJDI for True Positive Rate Disparities}
\label{sec:lambda-p}

The discussion in Section~\ref{sec:ijdi} focuses on IJDI for FPR disparities, for the subpopulation $D_N$ of negative individuals ($y_{i,0}=0$).  We now consider a corresponding definition of IJDI for true positive rate (or equivalently, false negative rate) disparities, for the subpopulation $D_P$ of positive individuals ($y_{i,0}=1$).  Letting $p(S) = \frac{1}{|S_P|} \sum_{i \in S_P} p_i$, $p({\sim} S) = \frac{1}{|D_P \setminus S_P|} \sum_{i \in D_P \setminus S_P} p_i$, $\bar p = \frac{1}{|D_P|} \sum_{i \in D_P} p_i$, $TPR(S) = \frac{1}{|S_P|} \sum_{i \in S_P} \hat p_{i,b}$, $TPR({\sim} S) = \frac{1}{|D_P \setminus S_P|} \sum_{i \in D_P \setminus S_P} \hat p_{i,b}$, and $\bar p_b = \frac{1}{|D_P|} \sum_{i \in D_P} \hat p_{i,b}$, we can define the IJDI fairness criterion for TPR as:
\begin{equation}\label{eqn:IJDI-TPR}
TPR(S) - TPR({\sim} S) \le \lambda_P(p(S) - p({\sim} S)),
\end{equation}
if $p(S) > p({\sim} S)$, and $TPR(S) \le TPR({\sim} S)$ otherwise, where $\lambda_P$ is a user-defined constant.
Appendix~\ref{sec:utility-pos} derives an equivalent, utility-based formulation of the IJDI fairness criterion for TPR, justifying the linear form of eqn.~(\ref{eqn:IJDI-TPR}) and providing an intuitive expression for $\lambda_P$.  
Appendix~\ref{sec:refactor-pos} demonstrates the equivalence of these two formulations.  Finally, the approach for eliciting $\lambda_N$, described in Appendix~\ref{sec:eliciting}, can be used with minor modifications to elicit $\lambda_P$ as well.

\section{Utility-Based Derivation of the IJDI Criteria and $\lambda$ Parameters}
\label{sec:utility}

\subsection{Negative IJDI criterion (for $D_N$) and $\lambda_N$ parameter}
\label{sec:utility-neg}

In this section, we justify why the IJDI criterion for FPR defined in Section~\ref{sec:ijdi} includes a linear function of the difference in base rates, starting from a general notion of utility. We also demonstrate how this utility-based approach provides specific guidance for choosing $\lambda_N$ based on the desired tolerance to group-level differences in FPR relative to the cost of individual false positive and false negative errors.
We begin by defining a more general notion of \emph{utility}, $U_i$, as the expected difference between $U_{i,1}$ and $U_{i,0}$, where $U_{i,1}$ is the utility of individual $i$ receiving binarized recommendation $\hat p_{i,b}=1$, 
and $U_{i,0}$ is the utility of individual $i$ receiving binarized recommendation $\hat p_{i,b} = 0$.  This definition can be further refined as follows:
\begin{align*}
U_i &= E[U_{i,1}-U_{i,0}] \\ 
&= p_i E[U_{i,1}-U_{i,0} \:|\: y_i=1] +(1-p_i)E[U_{i,1}-U_{i,0} \:|\: y_i=0] \\
&=E[U_{i,1}-U_{i,0}\:|\:y_i=0] + p_i(E[U_{i,1}-U_{i,0}\:|\:y_i=1] - E[U_{i,1}-U_{i,0}\:|\:y_i=0]) \\
&=-cost(FP)+p_i(cost(FN)+cost(FP)),
\end{align*}
where $cost(FP)$ and $cost(FN)$ are the utility costs of a false positive error and a false negative error respectively.  Since these costs are constant, we know $\bar U = \frac{1}{|D_N|} \sum_{i \in D_N} U_i = -cost(FP)+\bar p(cost(FN)+cost(FP))$, and thus $U_i - \bar U = (cost(FN)+cost(FP))(p_i - \bar p)$ for negative individuals $i\in D_N$, where $\bar p = \frac{1}{|D_N|} \sum_{i \in D_N} p_i$.
%Similarly, we know $\bar U_P = \frac{1}{|D_P|} \sum_{i \in D_P} U_i = -cost(FP)+\bar p_P(cost(FN)+cost(FP))$, and thus $U_i - \bar U_P = (cost(FN)+cost(FP))(p_i - \bar p_P)$ for $i\in D_P$. 

Now assume that there exists some intersectional subgroup $S$ with $FPR(S) > \bar p_b$, where $\bar p_b = \frac{1}{|D_N|} \sum_{i \in D_N} \hat p_{i,b}$ is the average FPR for all negative individuals. We consider the utility cost to society of this false positive rate disparity as: \[cost(\Delta_{FPR})|S_N|(FPR(S)-\bar p_b) = cost(\Delta_{FPR}) \sum_{i \in S_N}
(\hat p_{i,b} - \bar p_b).\]

Thus, if we wish to ensure that the societal cost of the false positive rate disparity for subgroup $S$ is outweighed by the utility gain, we can set:
\[
cost(\Delta_{FPR}) \sum_{i \in S_N}
(\hat p_{i,b} - \bar p_b) \le \sum_{i \in S_N}
(U_i - \bar U),
\]
and thus,
\[
\sum_{i \in S_N}
(\hat p_{i,b} - \bar p_b) \le 
\frac{cost(FP)+cost(FN)}{cost(\Delta_{FPR})} \sum_{i \in S_N} (p_i - \bar p).
\]
Thus we can write our IJDI fairness criterion for the negatives as:
\begin{align*}
\sum_{i \in S_N} (\hat p_{i,b} - \bar p_b) \le \lambda_N \sum_{i \in S_N} (p_i - \bar p),
\end{align*}
where
\begin{equation*}
\lambda_N = \frac{cost(FP)+cost(FN)}{cost(\Delta_{FPR})} 
 = \frac{1 + \frac{cost(FN)}{cost(FP)}}{\frac{cost(\Delta_{FPR})}{cost(FP)}},
\end{equation*}
as given in eqn.~(\ref{eqn:neg-from-utility}) of Section~\ref{sec:ijdi}. We demonstrate the equivalence of eqn.~(\ref{eqn:neg-from-utility}) and our original definition of IJDI for FPR (eqn. (\ref{eqn:IJDI-FPR})) in Appendix~\ref{sec:refactor-neg}.

Our expressions for the constants $\lambda_N$ and $\lambda_P$ utilize the societal costs of an individual false positive, an individual false negative, and an FPR or TPR disparity between groups. The ratio of $cost(FN)$ to $cost(FP)$ is commonly elicited in machine learning contexts such as cost-sensitive classification. In Appendix~\ref{sec:eliciting}, we discuss potential approaches to defining $cost(\Delta_{FPR})$ or $cost(\Delta_{TPR})$ based on our desired tolerance to group-level differences in error rate relative to the cost of individual false positive or false negative errors. 

The fact that the IJDI criterion uses a linear function of the difference in base rates is now clear, and arriving at this form from a general notion of utility is well-motivated under the assumption that error rate disparities are undesirable, creating a disutility proportional to the difference in error rates.

\subsection{Positive IJDI criterion (for $D_P$) and $\lambda_P$ parameter}
\label{sec:utility-pos}

In Appendix~\ref{sec:utility-neg} above, we show that the IJDI fairness criterion for false positive rate disparities can be written as:
\[
\sum_{i \in S_N} (\hat p_{i,b} - \bar p_b) \le \lambda_N \sum_{i \in S_N} (p_i - \bar p),
\]
where
\[
\lambda_N = \frac{cost(FP)+cost(FN)}{cost(\Delta_{FPR})} = \frac{1 + \frac{cost(FN)}{cost(FP)}}{\frac{cost(\Delta_{FPR})}{cost(FP)}}.
\]

We now derive the corresponding expression for true positive rate disparities, including an expression for the parameter $\lambda_P$, using our general, utility-based formulation.  Following the derivation in Appendix~\ref{sec:utility-neg}, we
know
$U_i - \bar U = (cost(FN)+cost(FP))(p_i - \bar p)$ for positive individuals $i\in D_P$, where 
$\bar U = \frac{1}{|D_P|} \sum_{i \in D_P} U_i$ and $\bar p = \frac{1}{|D_P|} \sum_{i \in D_P} p_i$.

Now assume that there exists some subgroup $S$ with $TPR(S) > \bar p_b$, where $\bar p_b = \frac{1}{|D_P|} \sum_{i \in D_P} \hat p_{i,b}$ is the average TPR for all positive individuals. We consider the utility cost to society of this TPR (or equivalently, FNR) disparity as: \[cost(\Delta_{TPR})|S_P|(TPR(S)-\bar p_b) = cost(\Delta_{TPR}) \sum_{i \in S_P}
(\hat p_{i,b} - \bar p_b).\]

Thus, if we wish to ensure that the societal cost of a true positive rate disparity for subgroup $S$ is outweighed by the utility gain, we can set:
\[
cost(\Delta_{TPR}) \sum_{i \in S_P}
(\hat p_{i,b} - \bar p_b) \le \sum_{i \in S_P}
(U_i - \bar U),
\]
and thus,
\[
\sum_{i \in S_P}
(\hat p_{i,b} - \bar p_b) \le 
\frac{cost(FP)+cost(FN)}{cost(\Delta_{TPR})} \sum_{i \in S_P} (p_i - \bar p).
\]
Thus we can write our IJDI fairness criterion for the positives as:
\begin{align}
\label{eqn:pos-from-utility}
\sum_{i \in S_P} (\hat p_{i,b} - \bar p_b) \le \lambda_P \sum_{i \in S_P} (p_i - \bar p),
\end{align}
where
\begin{equation*}
\lambda_P = \frac{cost(FP)+cost(FN)}{cost(\Delta_{TPR})}
= \frac{1+\frac{cost(FP)}{cost(FN)}}{\frac{cost(\Delta_{TPR})}{cost(FN)}}.
\end{equation*}

Again, the fact that the IJDI criterion uses a linear function of the disparity of base rates is now clear, and we also have derived an expression for the constant $\lambda_P$ in terms of the societal costs of an individual false positive, an individual false negative, and a TPR (or equivalently, FNR) disparity between groups. We demonstrate the equivalence of eqn.~(\ref{eqn:pos-from-utility}) and our original definition of IJDI for TPR (eqn. (\ref{eqn:IJDI-TPR})) in Appendix~\ref{sec:refactor-pos}.

\section{Eliciting the Relative Cost of Error Rate Disparities}
\label{sec:eliciting}

In this section, we consider how the ratios $\frac{cost(\Delta_{FPR})}{cost(FP)}$ and $\frac{cost(\Delta_{TPR})}{cost(FN)}$, might be elicited from a policy-maker, since estimates of these quantities are needed to compute the parameters $\lambda_N$ and $\lambda_P$ for the negative and positive IJDI criteria respectively.
Focusing on the negative IJDI criterion, our approach is to estimate the policy-maker's indifference curve between one potential scenario with unequal FPR across subgroups, and another potential scenario with equal FPR across subgroups but a higher overall FPR.

While full exploration of the preference elicitation problem is beyond the scope of this paper, we present an example of how this might be done for the COMPAS dataset.  We consider asking a policy-maker the following question: \\

``Imagine you have two subgroups of defendants, Group $A$ and Group $B$, differing in some demographic attribute (e.g., male vs. female, or Black vs. white).  You also have two automated systems, System 1 and System 2, which could be used to predict whether each defendant is ``high risk'' or ``low risk''.  You would like to avoid both false positives (predicting ``high risk'' for a defendant who does not reoffend) and false negatives (predicting ``low risk'' for a defendant who does reoffend), as well as balancing the rates of false positive and false negative errors between Group $A$ and Group $B$ (to avoid unfairly disadvantaging either group).

To test the two systems, you first run System 1 to predict risk (``high'' or ``low'') for a historical sample of 100 individuals from Group $A$ who did not reoffend.  You see that [$z_1$] of the 100 individuals were predicted as "high risk", corresponding to a [$z_1$]\% false positive rate for Group $A$. 

Similarly, you run System 1 to predict risk for a historical sample of 100 individuals from Group $B$ who did not reoffend.  You see that [$z_2$] of the 100 individuals were predicted as ``high risk'', corresponding to a [$z_2$]\% false positive rate for Group $B$, and a total false positive rate of $\left[\frac{z_1+z_2}{2}\right]$\%.

Next, you run System 2 for the same 200 individuals, and see that it returns the same false positive rate for both Group $A$ and Group $B$.  How high would this false positive rate have to be for you to prefer System 1 to System 2?

Your answer: [$z_3$]\%'' \\

To use this scenario in practice, we can draw integers $z_1$ and $z_2$
from a discrete uniform distribution on \{0, 1, 2, \ldots, 100\}, requiring $z_1 \ne z_2$ and $z_1 + z_2$ even, ask the policy-maker the question (filling in the chosen values of $z_1$ and $z_2$), and elicit their response $z_3$. If $z_3 < \frac{z_1 + z_2}{2}$, we warn the user that they are choosing a system with higher FPR and a larger FPR disparity, and allow them to re-answer the question.  Otherwise, we compute and record $(x_i, y_i) = \left(|z_1-z_2|, z_3-\frac{z_1+z_2}{2}\right)$, where $x_i$ denotes an FPR disparity between groups and $y_i$ denotes an overall FPR increase with equal disutility.

Then, to obtain $\frac{cost(\Delta_{FPR})} {cost(FP)}$ from a single question, 
we set the disutility from the FPR disparity,
$cost(\Delta_{FPR}) \times 100 \times \frac{x_i}{2}$,
equal to the disutility from the overall FPR increase, $cost(FP)\times 200 \times y_i$, and thus
\[\frac{cost(\Delta_{FPR})} {cost(FP)} = \frac{4y_i}{x_i}
 = \frac{4 z_3-2 z_1-2 z_2}{|z_1-z_2|}.
\]

Alternatively, we can ask the user 
multiple questions with different
values of $z_1$ and $z_2$, thus obtaining a set of $(x_i, y_i)$ pairs. We then fit the linear equation $y_i = \beta x_i$ by ordinary least squares linear regression, and set 
$\frac{cost(\Delta_{FPR})} {cost(FP)} = 4\beta$, obtaining
\[
\frac{cost(\Delta_{FPR})} {cost(FP)} = \frac{4 \sum_i x_i y_i}{\sum_i x_i^2},
\]
which is a weighted average of 
the values $\frac{4z_3-2z_1-2z_2}{|z_1-z_2|}$ with weights $|z_1-z_2|$.

Finally, while we have focused above on eliciting $\frac{cost(\Delta_{FPR})}{cost(FP)}$, we note that $\frac{cost(\Delta_{TPR})}{cost(FN)}$ can be elicited similarly by using the same scenario, but referring to the subpopulation who reoffended rather than the subpopulation who did not reoffend, and false negatives rather than false positives. 

\section{Equivalence of Different Formulations of the IJDI Fairness Criteria}
\label{sec:refactor}

\subsection{Equivalence for the negative IJDI criterion}
\label{sec:refactor-neg}

Here we show that our original definition of IJDI for false positive rate disparities (eqn.~(\ref{eqn:IJDI-FPR})) is equivalent to the definition which we derive from our utility-based formulation (eqn.~(\ref{eqn:neg-from-utility})).
As above, let $p(S) = \frac{1}{|S_N|} \sum_{i \in S_N} p_i$, $p({\sim} S) = \frac{1}{|D_N \setminus S_N|} \sum_{i \in D_N \setminus S_N} p_i$, $\bar p = \frac{1}{|D_N|} \sum_{i \in D_N} p_i$, $FPR(S) = \frac{1}{|S_N|} \sum_{i \in S_N} \hat p_{i,b}$, $FPR({\sim} S) = \frac{1}{|D_N \setminus S_N|} \sum_{i \in D_N \setminus S_N} \hat p_{i,b}$, and $\bar p_b = \frac{1}{|D_N|} \sum_{i \in D_N} \hat p_{i,b}$.  We consider the IJDI fairness criterion for FPR in eqn.~(\ref{eqn:IJDI-FPR}):
\begin{equation*}
FPR(S) - FPR({\sim} S) \le \lambda_N(p(S) - p({\sim} S)).
\end{equation*}

Then we can write \[FPR(S) - FPR({\sim} S) = \frac{|D_N|}{|S_N|}(FPR(S) - \bar p_b),\]
which follows from the fact that $|S_N| FPR(S) + |{\sim} S_N| FPR({\sim} S) = |D_N| \bar p_b = \sum_{i \in D_N} \hat p_{i,b}$.
Similarly, we can write
\[ p(S) - p({\sim} S) = 
\frac{|D_N|}{|S_N|}(p(S) - \bar p),\]
which follows from the fact that $|S_N| p(S) + |{\sim} S_N| p({\sim} S) = |D_N| \bar p = \sum_{i \in D_N} p_i$.  Plugging these into the IJDI fairness criterion for FPR, we obtain:
\[ \frac{|D_N|}{|S_N|}(FPR(S) - \bar p_b) \le \lambda_N \frac{|D_N|}{|S_N|}(p(S) - \bar p),\]
and thus,
\[ FPR(S) - \bar p_b \le \lambda_N (p(S) - \bar p). \]
Next, we plug in the definitions of $FPR(S)$ and $p(S)$, and multiply through by $|S_N|$, to obtain: 
\[ \left( \sum_{i \in S_N} \hat p_{i,b} \right) - |S_N| \bar p_b \le \lambda_N \left(\left( \sum_{i \in S_N} p_i \right) - |S_N| \bar p\right), \] 
and thus we obtain the expression in eqn.~(\ref{eqn:neg-from-utility}),
\[ \sum_{i \in S_N} (\hat p_{i,b} - \bar p_b) \le \lambda_N \sum_{i \in S_N} (p_i - \bar p). \]

\subsection{Equivalence for the positive IJDI criterion}
\label{sec:refactor-pos}

Here we show that our definition of IJDI for true positive rate disparities (eqn.~(\ref{eqn:IJDI-TPR})) is equivalent to the definition which we derive from our utility-based formulation (eqn.~(\ref{eqn:pos-from-utility})).
Given the IJDI criterion, $TPR(S) - TPR({\sim} S) \le \lambda_P(p(S) - p({\sim} S))$, we can first write \[TPR(S) - TPR({\sim} S) = \frac{|D_P|}{|S_P|}(TPR(S) - \bar p_b),\]
which follows from the fact that $|S_P| TPR(S) + |{\sim} S_P| TPR({\sim} S) = |D_P| \bar p_b = \sum_{i \in D_P} \hat p_{i,b}$.
Similarly, we can write
\[ p(S) - p({\sim} S) = 
\frac{|D_P|}{|S_P|}(p(S) - \bar p),\]
which follows from the fact that $|S_P| p(S) + |{\sim} S_P| p({\sim} S) = |D_P| \bar p = \sum_{i \in D_P} p_i$.  Plugging these into the IJDI fairness criterion for TPR, we obtain:
\[ \frac{|D_P|}{|S_P|}(TPR(S) - \bar p_b) \le \lambda_P \frac{|D_P|}{|S_P|}(p(S) - \bar p),\]
and thus,
\[ TPR(S) - \bar p_b \le \lambda_P (p(S) - \bar p). \]
Next, we plug in the definitions of $TPR(S)$ and $p(S)$, and multiply through by $|S_P|$, to obtain: 
\[ \left( \sum_{i \in S_P} \hat p_{i,b} \right) - |S_P| \bar p_b \le \lambda_P \left(\left( \sum_{i \in S_P} p_i \right) - |S_P| \bar p\right), \] 
and thus we obtain the expression in eqn.~(\ref{eqn:pos-from-utility}),
\[ \sum_{i \in S_P} (\hat p_{i,b} - \bar p_b) \le \lambda_P \sum_{i \in S_P} (p_i - \bar p). \] 

\section{Edge Cases}
\label{sec:edge-cases}
There are several edge cases that arise from the expressions pertaining to the IJDI fairness criterion and IJDI-Scan defined in Sections~\ref{sec:ijdi} and~\ref{sec:ijdi-scan}. We now discuss which conditions can be violated and how to make corrections when these cases occur.  We focus here on the negative IJDI-Scan, but note that these corrections are made for positive IJDI-Scan as well.

\subsection{Edge Case 1: $p(S) < p({\sim} S)$.} As noted above, our negative IJDI fairness criterion requires $FPR(S) - FPR({\sim} S) \le \lambda_N(p(S) - p({\sim} S))$ when $p(S) \ge p({\sim} S)$.  But when $p(S) < p({\sim} S)$, requiring an error rate imbalance $FPR(S) - FPR({\sim} S) \le \lambda_N(p(S) - p({\sim} S)) < 0$ favoring subgroup $S$ in order to avoid violation of the criterion would be too strict; rather, it simply requires $FPR(S) \le FPR({\sim} S)$ (i.e., that there is no error rate imbalance against subgroup $S$) to achieve fairness.

Thus, when running IJDI-Scan, we must check whether the initially detected subgroup $S$ -- the subgroup with the most significant IJDI -- has $p(S) < p({\sim} S)$. If so, we correct this edge case by increasing the probabilities $p_i$ for $i \in S$, so that $p(S) = p({\sim} S)$, and re-running the scan.  In order to preserve the relative ordering of $p_i$ values, our approach is to shift all individual values $p_i$, $i\in S$, with $p_i < p({\sim} S)$, toward $p({\sim} S)$ as follows:
$$p_i := p_i + \alpha(p({\sim} S)-p_i),$$ where $$\alpha = \frac{\sum_{i \in S}(p({\sim} S) - p_i)}{\sum_{i\in S:p_i<p({\sim} S)}(p({\sim} S)-p_i)}.$$ This approach guarantees that $p(S) = p({\sim} S)$ and therefore ensures that the edge case is no longer violated. Note that the probabilities $p_i$ are monotonically increased by this step, thus increasing $\hat{p}_{i,scan}$ and reducing the score of subgroup $S$.  See Appendix~\ref{sec:edge} for the derivation of $\alpha$.

\subsection{Edge Case 2: $\hat{p}_{i,scan}$ outside $[0,1]$.} In the new definition of $\hat{p}_{i,scan}$, large $\lambda$ combined with nonzero differences in base rates can yield $\hat{p}_{i,scan}<0$ or $\hat{p}_{i,scan}>1$. Since $\hat{p}_{i,scan}$ represents a probability and is compared to the binary outcome $y_{i,scan}$ in the hypothesis test of traditional Bias Scan, its value must fall between $0$ and $1$. To handle the case when $\hat{p}_{i,scan} = \bar p_b + \lambda(p_i - \bar p) < 0$, we set $\hat{p}_{i,scan} := 0$, and when $\hat{p}_{i,scan} > 1$, we set $\hat{p}_{i,scan} := 1$.  However, an additional complication arises due to this censoring approach: if censoring reduces $\hat p_{i,scan}$ it may result in detection of subgroups as ``significant'' that do not actually have IJDI.

Let us define the means of the censored and uncensored $\hat{p}_{i,scan}$ values in subgroup $S$ as follows: 
$$E[censored] = \frac{\sum_{i \in S} \hat{p}_{i, censored}}{|S|}, \: E[uncensored] = \frac{\sum_{i \in S} \hat{p}_{i, uncensored}}{|S|}.$$ 

When $E[censored] < E[uncensored]$ due to there being many values satisfying $\hat{p}_{i, uncensored} > 1$ prior to censoring, subsets that do not actually have IJDI may be found due to the scan underestimating $\hat{p}_{i, scan}$. In this case, we wish to increase the $\hat{p}_{i, uncensored}$ values in $S$ that are below $1$, prior to censoring, while still preserving their ordering. 

There are two cases of interest:
\begin{enumerate}
    \item If $E[uncensored] \ge 1$, then we can simply set $\hat{p}_{i, uncensored} := 1$ for all $i \in S$. This approach guarantees that $E[censored] = 1$, so when we re-run the scan it will no longer find this subset.
    \item If $E[uncensored] < 1$, then we need to increase $\hat{p}_{i, uncensored}$ so that $E[censored]$ after adjustment is equal to the original $E[uncensored]$. To achieve this, for $i \in S$ where $\hat{p}_{i, uncensored} < 1$, we set $$\hat{p}_{i, uncensored} := \hat{p}_{i, uncensored} + \beta(1-\hat{p}_{i, uncensored}),$$ where $$\beta = \frac{\sum_{i\in S:\hat{p}_{i, uncensored}\geq 1}(\hat{p}_{i, uncensored} - 1)}{\sum_{i\in S:\hat{p}_{i, uncensored}<1} (1-\hat{p}_{i, uncensored})}.$$ This approach guarantees that $E[censored] = E[uncensored]$. Note that the probabilities $\hat{p}_{i,scan}$ are monotonically increased by this step, thus reducing the score of subgroup $S$. See Appendix~\ref{sec:edge} for the derivation of $\beta$.
\end{enumerate}

\subsection{Deriving the edge case corrections}
\label{sec:edge}
In this section, we derive definitions of the constants $\alpha$ and $\beta$, which allow us to resolve the edge cases above by adjusting the probabilities $p_i$. First, when $p(S) < p({\sim} S)$ for the most significant (highest-scoring) subgroup $S$, we wish to increase the individual probabilities $p_i$ in $S$ so that $p(S)=p({\sim} S)$, while still preserving their ordering. To achieve this condition, we set 
$p_i := p_i + \alpha(p({\sim} S)-p_i)$ for all $i \in S$ with $p_i < p({\sim} S)$, where
\[ \alpha = \frac{\sum_{i \in S}(p({\sim} S) - p_i)}{\sum_{i\in S:p_i<p({\sim} S)}(p({\sim} S)-p_i)}. \]

To see that this choice of $\alpha$
ensures that $\frac{\sum_{i \in S}p_i}{|S|} = p({\sim} S)$ after adjustment, we can write:
\begin{align*}
\frac{\sum_{i \in S} p_{i,adjusted}}{|S|} 
&= \frac{\sum_{i\in S:p_i<p({\sim} S)}(p_i + \alpha(p({\sim} S)-p_i)) + \sum_{i\in S:p_i\ge p({\sim} S)}p_i}{|S|} \\
&= \frac{\sum_{i \in S} p_i}{|S|} + \alpha \left(\frac{\sum_{i\in S:p_i<p({\sim} S)}(p({\sim} S)-p_i)}{|S|}\right)  \\
&= \frac{\sum_{i \in S} p_i}{|S|} + \left( \frac{\sum_{i \in S}(p({\sim} S) - p_i)}{\sum_{i\in S:p_i<p({\sim} S)}(p({\sim} S)-p_i)} \right) \left( \frac{\sum_{i\in S:p_i<p({\sim} S)}(p({\sim} S)-p_i)}{|S|} \right)  \\
&= \frac{\sum_{i \in S} p_i}{|S|} + \frac{\sum_{i \in S}(p({\sim} S) - p_i)}{|S|}  \\
&= p({\sim} S).
\end{align*}

For the second edge case, when $\hat{p}_{i, scan}$ falls outside of $[0,1]$, we set $\hat{p}_{i, scan} := 1$ if $\hat{p}_{i, scan} > 1$ and $\hat{p}_{i, scan} := 0$ if $\hat{p}_{i, scan} < 0$. However, if this censoring step reduces $\hat p_{i,scan}$, it may result in detection of subgroups as ``significant'' that do not actually have IJDI. To address this, if $E[uncensored] \ge 1$, then we set $\hat{p}_{i, uncensored} := 1$ for all $i \in S$ to ensure that $E[censored] = 1$. If $E[uncensored] < 1$, then we increase $\hat{p}_{i, uncensored}$ so that $E[censored]$ after adjustment is equal to the original $E[uncensored]$. To achieve this condition, we set $\hat{p}_{i, uncensored} := \hat{p}_{i, uncensored} + \beta(1-\hat{p}_{i, uncensored})$ for all $i \in S$ with $\hat{p}_{i, uncensored} < 1,$ where
$$\beta = \frac{\sum_{i\in S:\hat{p}_{i, uncensored}\geq 1}(\hat{p}_{i, uncensored} - 1)}{\sum_{i\in S:\hat{p}_{i, uncensored}<1} (1-\hat{p}_{i, uncensored})}.$$ 
To see that this choice of $\beta$ ensures that $E[censored] = E[uncensored]$ after adjustment, we can write:
\begin{align*}
\frac{\sum_{i \in S} \hat{p}_{i, adjusted,censored}}{|S|}
&=
\frac{\sum_{i \in S: \hat{p}_{i, uncensored}<1} (\hat{p}_{i, uncensored}+ \beta(1-\hat{p}_{i, uncensored}))}{|S|} \\
&+ \frac{\sum_{i \in S: \hat{p}_{i, uncensored}\ge 1} 1}{|S|} \\ 
&= \frac{\sum_{i \in S} \hat{p}_{i,uncensored}}{|S|} +
\beta \frac{\sum_{i \in S: \hat{p}_{i, uncensored}<1} (1-\hat{p}_{i, uncensored})}{|S|} \\
&+ 
\frac{\sum_{i \in S: \hat{p}_{i, uncensored}\ge 1} (1 - \hat{p}_{i,uncensored})}{|S|} \\
&= \frac{\sum_{i \in S} \hat{p}_{i,uncensored}}{|S|} +
\frac{\sum_{i \in S: \hat{p}_{i, uncensored} \ge 1} (\hat{p}_{i, uncensored}-1)}{|S|} \\
&+ 
\frac{\sum_{i \in S: \hat{p}_{i, uncensored}\ge 1} (1 - \hat{p}_{i,uncensored})}{|S|} \\ 
&= \frac{\sum_{i \in S} \hat{p}_{i,uncensored}}{|S|}.
\end{align*}

\section{The IJDI-Scan Algorithm}
\label{sec:ijdi-algo}

Algorithm~\ref{alg:algorithm} details the implementation of the IJDI-Scan algorithm. 
The algorithm iteratively runs the scan (line 7),
checks if the subset found by the scan violates edge case 1 (line 8) or edge case 2 (line 10), and makes any required adjustments.  This process is repeated until 
no edge case conditions are violated. The algorithm is guaranteed to terminate based on the definitions of the edge case corrections (see Appendix~\ref{sec:edge-cases}), and in particular, the fact that both edge case corrections monotonically increase the values of $\hat p_{i,scan}$, both censored and uncensored. The boundedness and monotonicity of $\hat{p}_{i,censored}$ together imply convergence.

Algorithm~\ref{alg:algorithm} is first run on the original dataset 
($X$, $\hat{p}_b$, $p$, $y_0$) to obtain the subgroup $S^*$ with the most significant violation of our IJDI fairness criterion (i.e., the highest-scoring subgroup).  To compute the statistical significance of subgroup $S^*$, we perform \emph{randomization testing} by (i) generating a large number of synthetic datasets for which the null hypothesis $H_0$ holds (i.e., no insufficiently justified disparate impact is present), by redrawing the binarized recommendations $\hat{p}_{i,b}$ from the null distribution in eqn.~(\ref{ijdi-scan-hypothesis}); (ii) using Algorithm~\ref{alg:algorithm} to compute the maximum subgroup score for each null dataset; and (iii) comparing the score of the detected subgroup $S^*$ to the distribution of maximum subgroup scores under $H_0$.  The detected subgroup is statistically significant  at level $\alpha=.05$ if its score exceeds the 95th percentile of the distribution of maximum subgroup scores under $H_0$.

\begin{algorithm}[tb]
\caption{IJDI-Scan algorithm}
\label{alg:algorithm}
\textbf{Inputs}: features $\{x_i\}$, predictions $\{\hat{p}_{i,0}\}$, true probabilities $\{p_i\}$, outcomes $\{y_{i,0}\}$, thresholds $\{\theta_i\}$, parameter $\lambda$ \\
\textbf{Outputs}: Subgroup with most significant IJDI $S^*$, log-likelihood ratio score for IJDI $F(S^*)$
\begin{algorithmic}[1] %[1] enables line numbers
\STATE Filter dataset to only those records with $y_{i,0}=0$ (for FPR) or $y_{i,0}=1$ (for TPR).
\STATE violate\_flag := True
\STATE $\hat{p}_{i,b} := \mathbf{1}\{\hat{p}_{i,0} > \theta_i\}$, $\forall i$.
\WHILE{violate\_flag = True}
\STATE $\hat{p}_{i, uncensored} := \bar{p}_{b} + \lambda (p_i-\bar{p})$, $\forall i$.
\STATE $\hat{p}_{i, censored} :=\text{restrict}(\hat{p}_{i,uncensored}, [0,1])$, $\forall i$.
\STATE ($S^\ast$, $F(S^\ast)$) := bias\_scan($x_{scan} = \{x_i\}$, $\hat p_{scan} = \{\hat{p}_{i,censored}\}$, $y_{scan}=\{\hat p_{i,b}\}$)
\IF {$p(S^\ast) < p({\sim} S^\ast)$}
\STATE $p_{i} := p_{i} + \alpha (p({\sim} S^\ast)-p_{i})$ for $i \in S^\ast$, $p_{i} < p({\sim} S^\ast)$.
\ELSIF {$E[censored](S^\ast) < E[uncensored](S^\ast)$}
\STATE If $E[uncensored](S^\ast) \ge 1$, then $\hat{p}_{i, scan} := 1$ for $i \in S^\ast$, $\hat{p}_{i,scan} < 1$.
\STATE If $E[uncensored](S^\ast) < 1$, then $\hat{p}_{i, scan} := \hat{p}_{i, scan} + \beta (1-\hat{p}_{i, scan})$ for $i \in S^\ast$, $\hat{p}_{i,scan} < 1$.
\ELSE
\STATE violate\_flag := False
\ENDIF
\ENDWHILE
\STATE \textbf{return} ($S^\ast$, $F(S^\ast)$)
\end{algorithmic}
\end{algorithm}

\section{Description of Datasets}
\label{sec:datasets}

We conduct experiments using two public, widely used datasets: ProPublica's COMPAS dataset and the German Credit Dataset (Institut f\"ur Statistik und \"Okonometrie). Table~\ref{table:dataset-table} provides a summary of the two datasets. 

\begin{table}
  \begin{center}
    \begin{tabular}{c c c c} 

     \textbf{Dataset} & \textbf{$y_{i,0}=1$} & \textbf{$|X_m|$} & \textbf{$|D|$} \\ \hline
     COMPAS & Reoffended in 2 years & 5 & 7214 \\ 
     German Credit & Not Creditworthy & 9 & 1000  \\ 
    \end{tabular}
  \end{center}
  \caption{Summary of datasets used in our experiments. Binary outcome $y_{i,0}$, number of attributes $|X_m|$, and number of records $|D|$.}
  \label{table:dataset-table}
\end{table}

\subsection{COMPAS} We use the COMPAS (Correctional Offender Management Profiling for Alternative Sanctions) dataset, which has been made publicly available by ProPublica (\verb|https://github.com/propublica/compas-analysis/|). We use a binary outcome, which represents whether the defendant was re-arrested within two years, as our target variable. Further, we discretize the numerical features. To demonstrate the range of possible values each variable can take, the following subgroup represents the full dataset $D$: \{``sex'': [``Female'', ``Male''], ``race'': [``African-American'', ``Asian'',  ``Caucasian'', ``Hispanic'', ``Native American'' ``Other''], ``under 25'': [False, True], ``prior offenses’': [``None’', ``1 to 5’', ``Over 5’'], ``charge degree’': [``Misdemeanor’', ``Felony’']\}.  The COMPAS predictions (decile scores) are used only for the mitigation experiments in Sections~\ref{sec:mit-1} and~\ref{sec:mit-2}.  For these experiments, we define the predicted probabilities of reoffending $\hat{p}_{i,0}$ as the proportion of reoffenders among all defendants with the given decile score.  A threshold probability of 0.45, corresponding to a decile score of 5+ (``high risk''), is used to define $\hat{p}_{i,b}=1$.

\subsection{German Credit} We use the German Credit dataset, which originally comes from Professor Dr. Hans Hofmann of the Institut f\"ur Statistik und \"Okonometrie. This dataset has been made publicly available by the University of California, Irvine (\verb|https://archive.ics.uci.edu/ml/datasets/statlog+(german+credit+data|). We use the preprocessed version, available publicly on Kaggle (\verb|https://www.kaggle.com/datasets/kabure/german-credit-data-with-risk|), which provides a binary outcome variable representing whether a customer is creditworthy. We further discretize the data so that each variable is categorical. To demonstrate the range of possible values each variable can take, the following subgroup represents the full dataset $D$: \{``sex'': [``Female'', ``Male''], ``under 25'': [False, True], ``job’': [``None’', ``1’', ``2+’'], ``housing’': [``Free’', ``Rent’', ``Own''], ``savings’': [``N/A’', ``Little’', ``Moderate’', ``Quite Rich’', ``Rich’'], ``checking’': [``N/A’', ``Little’', ``Moderate’', ``Rich’'], ``credit amount’': [``Low’', ``Moderate’', ``High'', ``Very High''], ``duration’': [``Very Short’', ``Short’', ``Moderate’', ``Long’', ``Very Long’'], ``purpose’': [``Business’', ``Car’', ``Education’', ``Radio/TV’', ``Furniture/Equipment’', ``Domestic Appliances’', ``Repairs’', ``Vacation/Other’']\}.

\section{Deriving the $\lambda^*$ Cutoff Parameter for Experiment 1}
\label{sec:theoretical}
Recall from the first experiment of Section~\ref{sec:evaluation} that we simulate data by drawing predictions and outcomes using simple uniform distributions. In this section, we derive a closed form expression for the cutoff value $\lambda^\ast$, such that when $\lambda < \lambda^\ast$, we expect $S$ and ${\sim} S$ to have an error rate imbalance that is insufficiently justified by the base rate difference, and when $\lambda > \lambda^\ast$ we expect $S$ and ${\sim} S$ to have an error rate imbalance that is sufficiently justified (no IJDI).

We begin by deriving a normalized probability density function for the negative outcomes and the positive outcomes. Given a uniform random variable $X \:{\sim}\: [c \pm 0.01k],$ its probability density is $f(x)=\frac{1}{0.02k}, x \in [c \pm 0.01k]$. Given a Bernoulli random variable $Y \:{\sim}\: \text{Bernoulli}(p),$ $\Pr(Y=0)=1-p$ and $\Pr(Y=1)=p$. Then, we have:
$$p_N^{\ast}(x)=\frac{1-x}{0.02k}, p_P^{\ast}(x)=\frac{x}{0.02k}, x \in [c \pm 0.01k], k>0,$$
where $p_N^{\ast}(x)$ is the non-normalized distribution of $X$ given that Y is a negative ($Y=0$) and $p_P^{\ast}(x)$ is the non-normalized distribution of $X$ given that Y is a positive ($Y=1$). To normalize these distributions and achieve conditional pdfs $p_N(x)=\Pr(X=x\:|\:Y=0)$ and $p_P(x)=\Pr(X=x\:|\:Y=1)$, we can integrate over the domain in each case and divide the non-normalized distribution by this result. Integrating these functions yields:
$$\int_{c-0.01k}^{c+0.01k}p_N^{\ast}(x)dx = \int_{c-0.01k}^{c+0.01k}\frac{1-x}{0.02k}dx=1-c,$$
$$ \int_{c-0.01k}^{c+0.01k}p_P^{\ast}(x)dx = \int_{c-0.01k}^{c+0.01k}\frac{x}{0.02k}dx=c.$$
Thus, the conditional pdfs are
\begin{equation}
\begin{split}
p_N(x)=\frac{1-x}{0.02k(1-c)}&, p_P(x)=\frac{x}{0.02kc},  x \in [c \pm 0.01k], k>0.
\end{split}
\label{eqn:pdfs}
\end{equation}

Now we return to the setting of Experiment 1, where we draw predictions from uniform distributions such that $p_i \:{\sim}\: \text{Uniform}[0.51 \pm 0.01k]$ for $i\in S$ and $p_i \:{\sim}\: \text{Uniform}[0.49 \pm 0.01k]$ for $i\not\in S$, we assume a sharp threshold of $\theta=0.5$ for recommendations such that $\hat p_{i,b} = \mathbf{1}\{\hat p_{i,0} > 0.5\}$, and we simulate outcomes such that $y_{i,0} \:{\sim}\: \text{Bernoulli}(p_i)$.

First, let us find $\lambda_N^{\ast}$. Given the negative IJDI fairness criterion $FPR(S) - FPR({\sim} S) \le \lambda_N (p(S)-p({\sim} S))$, the theoretical cutoff for $\lambda_N^{\ast}$ is
$$\lambda_N^{\ast}=\frac{FPR(S) - FPR({\sim} S)}{p(S)-p({\sim} S)},$$
as any $\lambda < \lambda_N$ will result in IJDI (the fairness criterion is violated) and any $\lambda > \lambda_N$ will result in no IJDI (the fairness criterion holds). We now examine the cases where $0 \le k \le 1$ and $k \ge 1$ separately.

When $0 \le k \le 1$, we know based on how the probability distributions are centered that $p(S)=E(p_i \:|\: i \in S)=0.51$ and $p({\sim} S)=E(p_i \:|\: i \not\in S)=0.49$. We also know based on the domains of the distributions that $p_i > 0.5, \forall i \in S$ and $p_i < 0.5, \forall i \not\in S$, which yields $FPR(S)=P(p_i > 0.5 \:|\: i \in S)=1$ and $FPR({\sim} S)=P(p_i > 0.5 \:|\: i \not\in S)=0$. Thus, we have a straightforward calculation of $\lambda_N^{\ast}$:
$$\lambda_N^{\ast}=\frac{1-0}{0.51-0.49}=50.$$
When $k > 1$, we can use the pdfs from eqn.~(\ref{eqn:pdfs}) to calculate $\lambda_N^{\ast}$. We first calculate conditional expectation $E(X=x\:|\:Y=0)$:
\begin{align*}
    E(X=x \:|\: Y=0) &= \int_{c-0.01k}^{c+0.01k}xp_N(x)dx \\
    &= \int_{c-0.01k}^{c+0.01k}x\frac{1-x}{0.02k(1-c)}dx = c - \frac{0.0001k^2}{3(1-c)}=A_N(c,k).
\end{align*}
Next, we calculate conditional probability $P(X>0.5 \:|\: Y=0)$:
\begin{align*}
    P(X>0.5\:|\:Y=0) &= \int_{0.5}^{c+0.01k}p_N(x)dx \\
    &= \int_{0.5}^{c+0.01k}\frac{1-x}{0.02k(1-c)}dx = \frac{6.25}{k(1-c)}-\frac{25(1-c)}{k} +\frac{1}{2}-\frac{0.0025k}{1-c}=B_N(c,k).
\end{align*}
Now, substituting into the expression for $\lambda_N^{\ast}$ yields:
\begin{align*}
    \lambda_N^{\ast} &=\frac{FPR(S) - FPR({\sim} S)}{p(S)-p({\sim} S)} \\
    & = \frac{E(p_i \:|\: i \in S) - E(p_i \:|\: i \not\in S)}{P(p_i > 0.5 \:|\: i \in S) - P(p_i > 0.5 \:|\: i \not\in S)} \\
    & = \frac{B_N(c,k)\rvert_{c=0.51}- B_N(c,k)\rvert_{c=0.49}}{A_N(c,k)\rvert_{c=0.51} - A_N(c,k)\rvert_{c=0.49}} \\
    & = \frac{75}{k}\frac{4999-k^2}{7497-k^2}.
\end{align*}

Now, using a similar procedure, let us find $\lambda_P^{\ast}$. Given the positive IJDI fairness criterion $TPR(S) - TPR({\sim} S) \le \lambda_P (p(S)-p({\sim} S))$, the theoretical cutoff for $\lambda_P^{\ast}$ is
$$\lambda_P^{\ast}=\frac{TPR(S) - TPR({\sim} S)}{p(S)-p({\sim} S)},$$
as any $\lambda < \lambda_P$ will result in IJDI and any $\lambda > \lambda_P$ will result in no IJDI. We now examine the cases where $0 \le k \le 1$ and $k \ge 1$ separately.

When $0 \le k \le 1$, just as was the case for the negatives, we know that $p(S)=E(p_i
\:|\:i \in S)=0.51$, $p({\sim} S)=E(p_i\:|\:i \not\in S)=0.49$, $TPR(S)=P(p_i > 0.5 \:|\: i \in S)=1$, and $TPR({\sim} S)=P(p_i > 0.5 \:|\: i \not\in S)=0$. Thus, we have the same straightforward calculation of $\lambda_P^{\ast}$:
$$\lambda_P^{\ast}=\frac{1-0}{0.51-0.49}=50.$$
When $k > 1$, we can use the pdfs from eqn.~(\ref{eqn:pdfs}) to calculate $\lambda_P^{\ast}$. We first calculate conditional expectation $E(X=x\:|\:Y=1)$:
\begin{align*}
    E(X=x\:|\:Y=1) &= \int_{c-0.01k}^{c+0.01k}xp_P(x)dx \\
    &= \int_{c-0.01k}^{c+0.01k}x\frac{x}{0.02kc}dx = c + \frac{0.0001k^2}{3c}=A_P(c,k).
\end{align*}
Next, we calculate conditional probability $P(X>0.5\:|\:Y=1)$:
\begin{align*}
    P(X>0.5\:|\:Y=1) &= \int_{0.5}^{c+0.01k}p_P(x)dx \\
    &= \int_{0.5}^{c+0.01k}\frac{x}{0.02kc}dx = \frac{25c}{k}+\frac{1}{2} +\frac{k}{400c}-\frac{6.25}{kc}=B_P(c,k).
\end{align*}
Now, substituting into the expression for $\lambda_P^{\ast}$ yields:
\begin{align*}
    \lambda_P^{\ast} &=\frac{TPR(S) - TPR({\sim} S)}{p(S)-p({\sim} S)} \\
    & = \frac{E(p_i\:|\: i \in S) - E(p_i\:|\: i \not\in S)}{P(p_i > 0.5\:|\: i \in S) - P(p_i > 0.5\:|\: i \not\in S)} \\
    & = \frac{B_P(c,k)\rvert_{c=0.51}- B_P(c,k)\rvert_{c=0.49}}{A_P(c,k)\rvert_{c=0.51} - A_P(c,k)\rvert_{c=0.49}} \\
    & = \frac{75}{k}\frac{4999-k^2}{7497-k^2}.
\end{align*}

Therefore, the $\lambda^*$ for both the negative IJDI-Scan (on $D_N$) and positive IJDI-Scan (on $D_P$) is:
$$\lambda^*_N = \lambda^*_P = 
\begin{cases}
    50 & 0 \leq k < 1 \\
    \frac{75}{k}\frac{4999-k^2}{7497-k^2} & k \geq 1 \\
\end{cases}$$

\section{Deriving a Log-Odds Shift}
\label{sec:shift}
Recall from Experiment 2 of Section~\ref{sec:experiments} that we set $logit(\hat{p}_{i,0}):=logit(p_i) + \gamma$ for $i\in S$, where $\gamma$ represents a log-odds shift. In this section, we derive an equivalent expression for directly setting $\hat{p}_{i,0}$ to accomplish this log-odds shift. We begin by taking the inverse of the $logit$ function. Let $\phi(p) = logit(p) = ln\Big(\frac{p}{1-p}\Big)$, where $p$ represents some probability. We can take the inverse to get an expression for converting from log-odds to probability:
\begin{align*}
& \phi(p) = ln\Big(\frac{p}{1-p}\Big) \\
\Longrightarrow \: & e^{\phi} = \frac{p}{1-p} \\
\Longrightarrow \: & p = \frac{e^{\phi}}{e^{\phi}+1}.
\end{align*}
Using this expression we can convert the log-odds transformation into a probability transformation:
\begin{align*}
    & \phi(\hat{p}_{i,0}):=\phi(p_i) + \gamma \\
    \Longrightarrow \: & \hat{p}_{i,0} := \frac{e^{(\phi(p_i) + \gamma)}}{e^{(\phi(p_i) + \gamma)} + 1} \\
    \Longrightarrow \: & \hat{p}_{i,0} := \frac{e^{ln\frac{p_i}{1-p_i}}e^{\gamma}}{e^{ln\frac{p_i}{1-p_i}}e^{\gamma} + 1} \\
    \Longrightarrow \: & \hat{p}_{i,0} := \frac{\frac{p_i}{1-p_i}e^{\gamma}}{\frac{p_i}{1-p_i}e^{\gamma} + 1} \\
    \Longrightarrow \: & \hat{p}_{i,0} := \frac{p_i e^{\gamma}}{p_i e^{\gamma} + 1 - p_i}.
\end{align*}

\section{Experimental Results for Positive IJDI-Scan}
\label{sec:pos-experiments}
\begin{figure*}[t]
\centering
(a) \includegraphics[width=0.4\columnwidth]{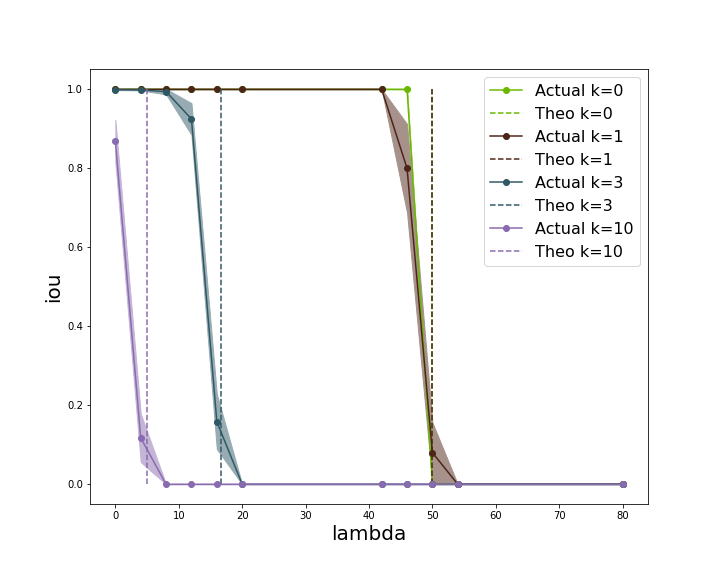}
(b) \includegraphics[width=0.4\columnwidth]{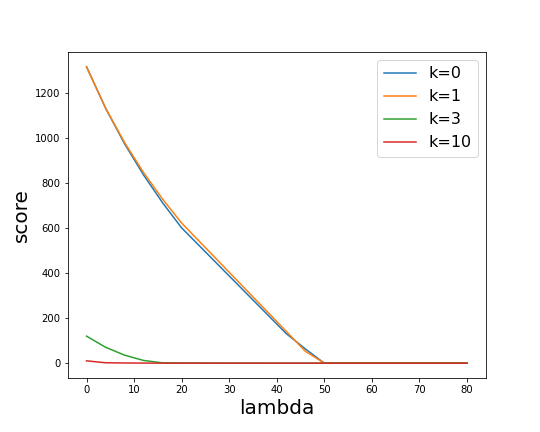}
\caption{Experiment 1 results for positive IJDI-Scan. (a) IOU (with 95\% CI) as a function of $\lambda$. For $k \in \{0,1,3,10\}$, empirical IOU values are plotted and the theoretical cutoff $\lambda^*$ is shown as a vertical dotted line. (b) Score as a function of $\lambda$, for $k \in \{0,1,3,10\}$.}
\label{fig:sim_1_pos}
\end{figure*}
\begin{figure}[t]
\centering
(a) \includegraphics[width=0.4\columnwidth]{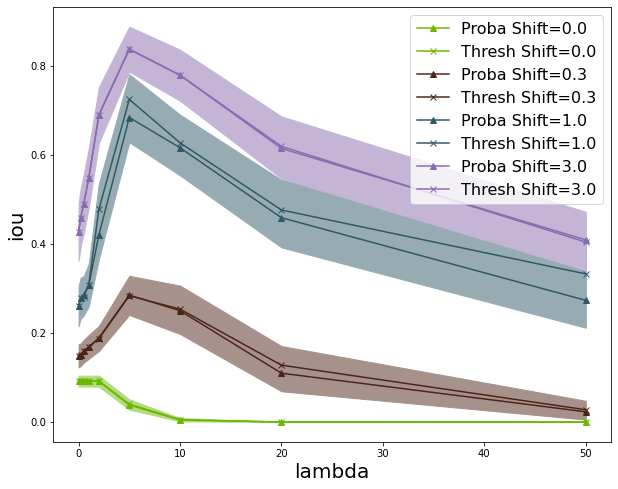}
(b) \includegraphics[width=0.4\columnwidth]{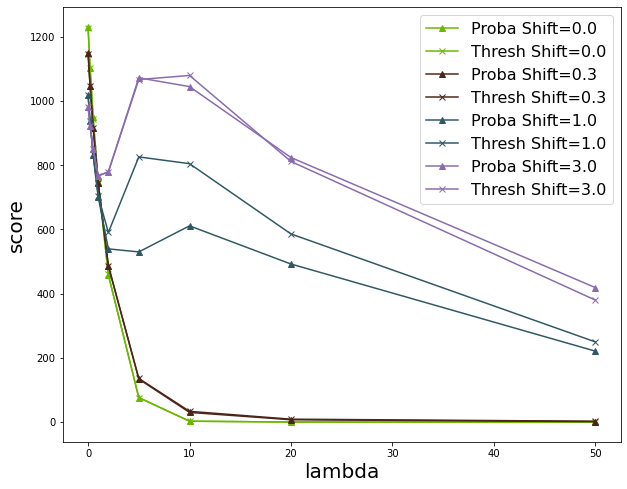}
\caption{Experiment 2 results for positive IJDI-Scan. (a) IOU (with 95\% CI) as a function of $\lambda$. For $\gamma \in \{0,0.3, 1, 3\}$, IOU values are plotted for an upward log-odds shift of probability (triangle) and downward log-odds shift of threshold (x). (b) Score as a function of $\lambda$, for $\gamma \in \{0,0.3, 1, 3\}$.}
\label{fig:sim_2_pos}
\end{figure}
\begin{figure}[t]
\centering
(a) \includegraphics[width=0.4\columnwidth]{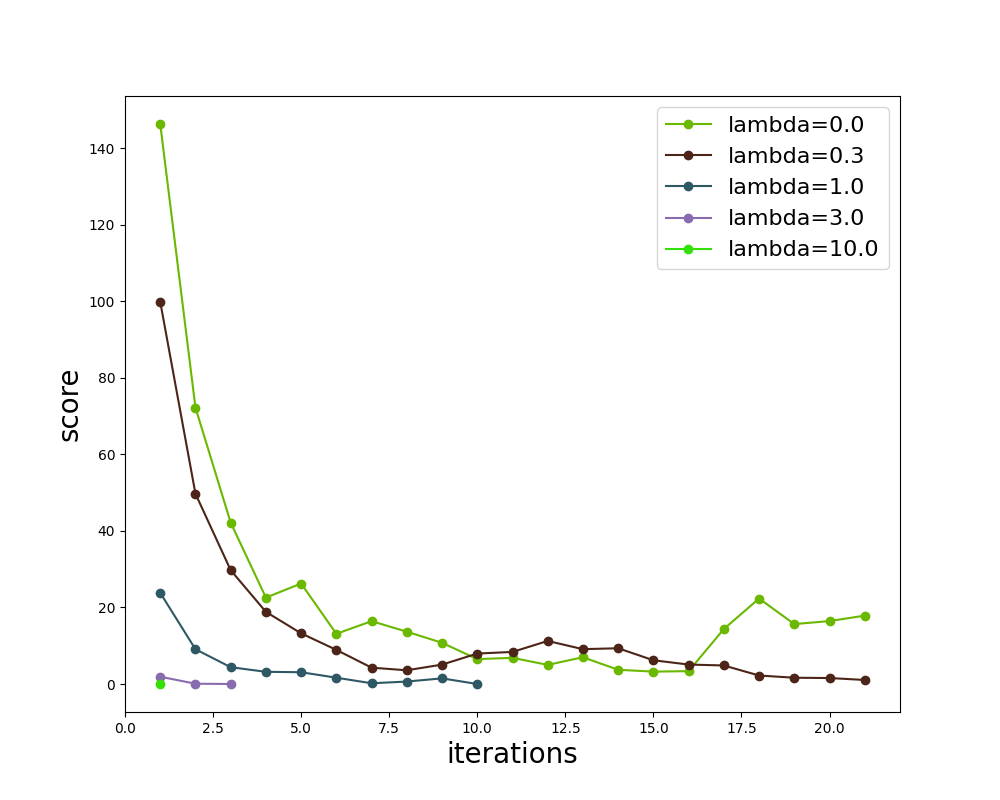}
(b) \includegraphics[width=0.4\columnwidth]{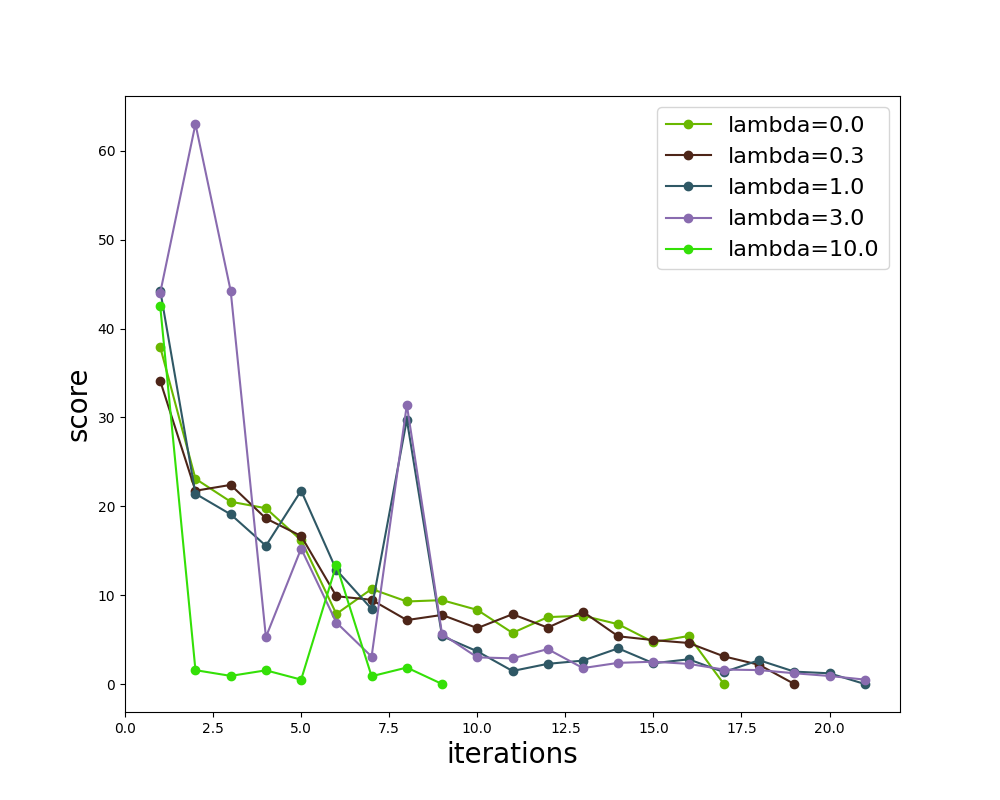}
\caption{Mitigation approach 2 results for positive IJDI-Scan. Score on (a) COMPAS and (b) German credit, as a function of the number of corrections.}
\label{fig:score_mit_2_pos}
\end{figure}

In this section we provide additional experimental results for the positive IJDI-Scan, including results from Experiment 1 (Figure~\ref{fig:sim_1_pos}), Experiment 2 (Figure~\ref{fig:sim_2_pos}), and Mitigation Approach 2 (Figure~\ref{fig:score_mit_2_pos}).  These results are consistent with the negative IJDI-Scan results above, demonstrating that positive IJDI-Scan can detect and mitigate disparities in true positive rates which are insufficiently justified by differences in base rates, whether these disparities result either from sharp thresholding of (correctly calibrated) probabilities or from systematic biases in the probabilities themselves.

One interesting finding in Figures~\ref{fig:sim_2_neg}(b) and~\ref{fig:sim_2_pos}(b) was 
that, while the score $F(S^\ast)$ generally decreases monotonically with $\lambda$, we observe some non-monotonicity in scores for both $\gamma=1$ and $\gamma=3$. This can result from our edge case correction process, in which probabilities are censored to the interval [0,1].  Our correction of the bias resulting from this censoring process restores $E[censored] = E[uncensored]$, but individual low probabilities may remain after censoring.  Larger $\lambda$ values are more likely to have such extreme probabilities, sometimes leading to higher scores.  Further investigation of this case will be addressed in future work.

\begin{figure}[t]
\centering
(a) \includegraphics[width=0.4\columnwidth]{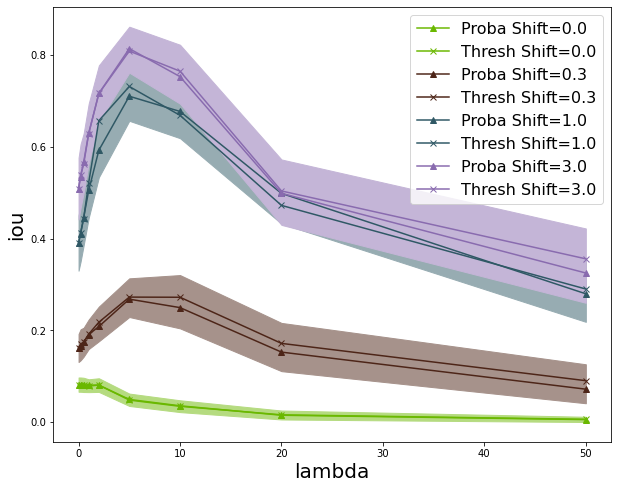}
(b) \includegraphics[width=0.4\columnwidth]{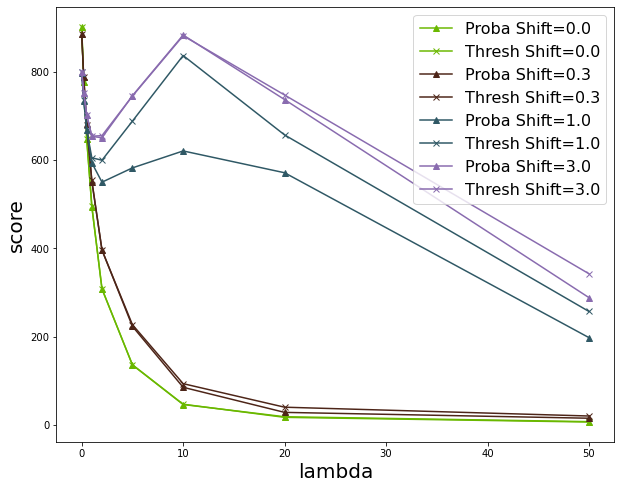}
\caption{Experiment 2 results for negative IJDI-Scan when using learned predictions for $p_i$. (a) IOU (with 95\% CI) as a function of $\lambda$. For $\gamma \in \{0,0.3, 1, 3\}$, IOU values are plotted for an upward log-odds shift of probability (triangle) and downward log-odds shift of threshold (x). (b) Score as a function of $\lambda$, for $\gamma \in \{0,0.3, 1, 3\}$.}
\label{fig:sim_2_neg_variant}
\end{figure}
\begin{figure}[t]
\centering
(a) \includegraphics[width=0.4\columnwidth]{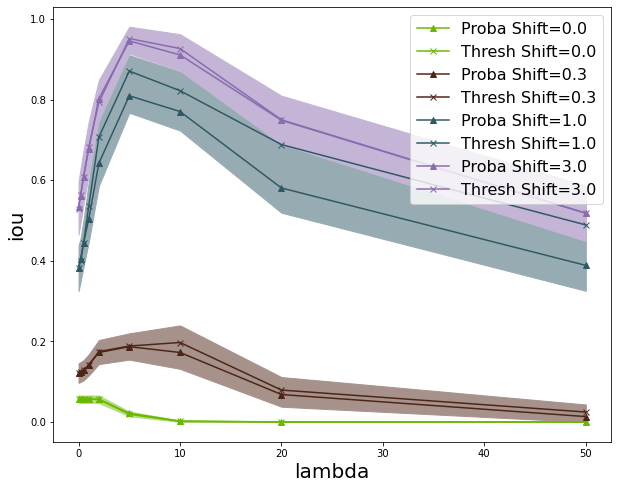}
(b) \includegraphics[width=0.4\columnwidth]{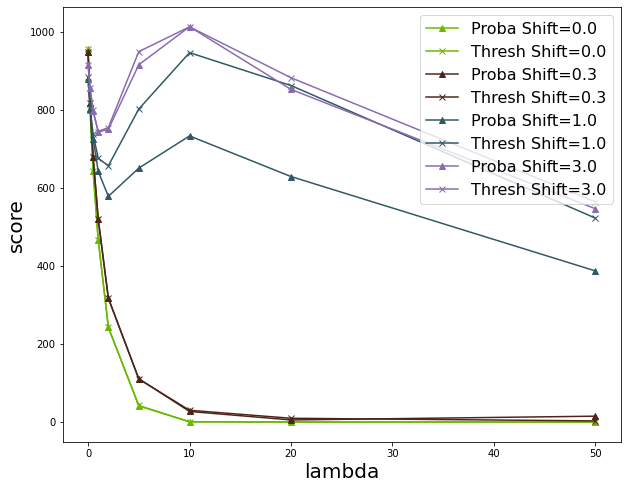}
\caption{Experiment 2 results for positive IJDI-Scan when using learned predictions for $p_i$. (a) IOU (with 95\% CI) as a function of $\lambda$. For $\gamma \in \{0,0.3, 1, 3\}$, IOU values are plotted for an upward log-odds shift of probability (triangle) and downward log-odds shift of threshold (x). (b) Score as a function of $\lambda$, for $\gamma \in \{0,0.3, 1, 3\}$.}
\label{fig:sim_2_pos_variant}
\end{figure}

\section{Experimental Results for Experiment 2 with Learned Probabilities}
\label{sec:exp-2-learned}
In this section we provide additional experimental results for Experiment 2, described in Section~\ref{sec:experiments}, when IJDI-Scan is run using learned rather than true probabilities. After generating the data using the true probabilities $p_i$ (as in Section~\ref{sec:experiments}), we build a logistic regression model on dataset $D$, predicting the outcomes $y_{i,0}$ using features $X_{i,1} \ldots X_{i,M}$, to generate new estimates of $p_i$. We then run the scan using these learned probabilities in place of the true probabilities $p_i$, for both the negative IJDI-Scan (Figure~\ref{fig:sim_2_neg_variant}) and positive IJDI-Scan (Figure~\ref{fig:sim_2_pos_variant}).  Comparing Figure~\ref{fig:sim_2_neg_variant} to Figure~\ref{fig:sim_2_neg}, and comparing Figure~\ref{fig:sim_2_pos_variant} to Figure~\ref{fig:sim_2_pos}, we see that running IJDI-Scan with learned predictions for $p_i$ yields results that are consistent with using true probabilities for $p_i$.  Detection power is very similar, as are the scores of the detected subgroups, and we continue to observe that IJDI-Scan with intermediate values of $\lambda$ outperforms both error rate balance ($\lambda = 0$) and risk-adjusted regression (large $\lambda$).

\section{Reporting Subgroups for Mitigation Approach 2}
\label{sec:report-subgroups}
In this section, we report the subgroups detected and mitigated by the initial run of IJDI-Scan as well as the subgroups detected and mitigated after the first and second corrections, for each value of $\lambda$.  The results of our mitigation runs on the COMPAS dataset are shown in Tables~\ref{table:compas-subset-neg-part1} through~\ref{table:compas-subset-pos-part2} for the negative and positive IJDI-Scans respectively.

For the negative scans with $\lambda_N = 0$ (equivalent to FPR-Scan) and small $\lambda_N =0.3$, the three detected high-FPR subgroups corresponded roughly to defendants with more than 5 prior offenses, defendants under 25 years old, and older African-American and Native American individuals respectively.  All of these clusters have both high FPR and high base rates of reoffending.
For larger $\lambda_N = 1$, the three detected high-FPR subgroups corresponded roughly to African-American and Hispanic defendants with more than 5 prior offenses facing felony charges, females under 25 facing felony charges, and African-American defendants.  
The second mitigated cluster for this case is particularly worth noting, as it has less of a false positive rate disparity than some of the other groups, but also a low base rate difference, thus making this subgroup more significant than other subgroups once we correct for base rate differences.  For even larger $\lambda_N \ge 3$, no clusters of significantly increased IJDI were found, suggesting that all false positive rate disparities would be justified for the given tradeoff between error rate balance and utility.

For the positive scans with $\lambda_P = 0$ (equivalent to TPR-Scan) and small $\lambda_P=0.3$, the three detected high-TPR subgroups corresponded roughly to African-American and Native American defendants with over 5 prior offenses, defendants under 25 years old, and defendants with over 5 prior offenses facing felony charges.  All of these clusters have both high TPR and high base rates of reoffending.  For larger $\lambda_P = 1$, the first two mitigated clusters are similar to the smaller $\lambda_P$ values, but the third cluster (white females under the age of 25) has both a relatively high TPR and a low base rate of reoffending.    Again, no clusters of significantly increased IJDI were identified for $\lambda_P \ge 3$.  These results demonstrate the ability of IJDI-Scan (with moderate values of $\lambda$) to pick up subtle intersectional biases (e.g., against younger females), which neither error rate balance ($\lambda=0$) nor risk-adjusted regression ($\lambda \gg 0$) would detect.

The results of our mitigation runs on the German credit dataset are shown in Tables~\ref{table:german-credit-subset-neg-part1} through~\ref{table:german-credit-subset-pos-part2}.  For this dataset. many of the IJDI-Scans (for both negative and positive individuals, across the various values of $\lambda$) detect similar subpopulations with characteristics that correlate to a high rate of non-creditworthiness, including lower checking and savings, longer loan durations, and either low or very high credit amounts.  It may be that the credit scoring agencies decide on individuals' creditworthiness based on sharp cutoffs on many of these characteristics, leading to large FPR and TPR disparities which persist even for high values of $\lambda$.

\clearpage
\begin{table*}
  \begin{center}
    \begin{tabular}{cc|c|ccccc} 
     $\lambda$ & Corrections & Score & Sex & Race \\
     \hline
     0 & 0 & 135.7 & all & all $\setminus$ [Other] \\
     0 & 1 & 83.3 & all & all $\setminus$ [Other] \\
     0 & 2 & 47.9 & all & [African-Am., Native Am.] \\
     0.3 & 0 & 74.5 & all & all $\setminus$ [Caucasian, Other] \\
     0.3 & 1 & 47.6 & all & all $\setminus$ [Native Am., Asian] \\
     0.3 & 2 & 27.1 & all & [African-Am., Native Am.] \\
     1 & 0 & 9.2 & all & [African-Am., Hispanic, Native Am.] \\
     1 & 1 & 7.5 & Female & [African-Am., Hispanic, Caucasian] \\
     1 & 2 & 10.4 & all & African-Am. \\
     3 & 0 & 0 & N/A & N/A \\
     10 & 0 & 0 & N/A & N/A \\
    \end{tabular}
  \end{center}
  \caption[Descriptions of biased negative subgroups for COMPAS (Part 1).]{Descriptions of the first three biased subgroups (with significantly high FPR after adjusting for base rates) detected and mitigated in the COMPAS dataset by negative IJDI-Scan [PART 1 OF 2]. The subset is shown for each value of $\lambda \in \{0,0.3,1,3,10\}$ and each number of previous corrections (0, 1, or 2). N/A represents that no subset was found by IJDI-Scan for $\lambda \ge 3$. Abbreviation: ``Am." = ``American".}
  \label{table:compas-subset-neg-part1}
\end{table*}

\begin{table*}
  \begin{center}
    \begin{tabular}{cc|ccccc} 
     $\lambda$ & Corrections & Under 25 & Prior Offenses & Charge Degree \\
     \hline
     0 & 0 & all & Over 5  & all \\
     0 & 1 & True & all & all \\
     0 & 2 & False & [1 to 5, Over 5] & all \\
     0.3 & 0 & all & Over 5  & all \\
     0.3 & 1 & True & all & all \\
     0.3 & 2 & False & 1 to 5 & all \\
     1 & 0 & all & Over 5  & Felony \\
     1 & 1 & True & [None, Over 5] & Felony \\
     1 & 2 & all & [None, 1 to 5] & all \\
     3 & 0 & N/A & N/A & N/A \\
     10 & 0 & N/A & N/A  & N/A \\
    \end{tabular}
  \end{center}
  \caption[Descriptions of biased negative subgroups for COMPAS (Part 2).]{Descriptions of the first three biased subgroups (with significantly high FPR after adjusting for base rates) detected and mitigated in the COMPAS dataset by negative IJDI-Scan [PART 2 OF 2]. The subset is shown for each value of $\lambda \in \{0,0.3,1,3,10\}$ and each number of previous corrections (0, 1, or 2). N/A represents that no subset was found by IJDI-Scan for $\lambda \ge 3$.}
  \label{table:compas-subset-neg-part2}
\end{table*}

\clearpage
\begin{table*}
  \begin{center}
    \begin{tabular}{cc|c|ccccc} 
     $\lambda$ & Corrections & Score & Sex & Race \\
     \hline
     0 & 0 & 146.3 & all & [African-Am., Native Am.] \\
     0 & 1 & 72.2 & all & all $\setminus$ [Other] \\
     0 & 2 & 42.2 & all & all $\setminus$  [African-Am., Asian] \\
     0.3 & 0 & 39.7 & all & [African-Am., Native Am.] \\
     0.3 & 1 & 27.9 & all & all $\setminus$ [Other] \\
     0.3 & 2 & 24.0 & all & all $\setminus$  [African-Am., Asian] \\
     1 & 0 & 23.8 & all & [African-Am., Native Am.] \\
     1 & 1 & 9.1 & all & all $\setminus$ [Caucasian, Other] \\
     1 & 2 & 4.4 & Female & Caucasian \\
     3 & 0 & 0 & N/A & N/A \\
     10 & 0 & 0 & N/A & N/A \\
    \end{tabular}
  \end{center}
  \caption[Descriptions of biased positive subgroups for COMPAS (Part 1).]{Descriptions of the first three biased subgroups (with significantly high TPR after adjusting for base rates) detected and mitigated in the COMPAS dataset by positive IJDI-Scan [PART 1 OF 2]. The subset is shown for each value of $\lambda \in \{0,0.3,1,3,10\}$ and each number of previous corrections (0, 1, or 2). N/A represents that no subset was found by IJDI-Scan for $\lambda \ge 3$. Abbreviation: ``Am." = ``American".}
  \label{table:compas-subset-pos-part1}
\end{table*}

\begin{table*}
  \begin{center}
    \begin{tabular}{cc|ccccc} 
     $\lambda$ & Corrections & Under 25 & Prior Offenses & Charge Degree \\
     \hline
     0 & 0 & all & Over 5  & all \\
     0 & 1 & True & [1 to 5, Over 5] & all \\
     0 & 2 & all & Over 5 & Felony \\
     0.3 & 0 & all & Over 5  & all \\
     0.3 & 1 & True & [1 to 5, Over 5] & all \\
     0.3 & 2 & all & Over 5 & Felony \\
     1 & 0 & all & Over 5  & all \\
     1 & 1 & True & 1 to 5 & all \\
     1 & 2 & True & all & all \\
     3 & 0 & N/A & N/A & N/A \\
     10 & 0 & N/A & N/A  & N/A \\
    \end{tabular}
  \end{center}
  \caption[Descriptions of biased positive subgroups for COMPAS (Part 2).]{Descriptions of the first three biased subgroups (with significantly high TPR after adjusting for base rates) detected and mitigated in the COMPAS dataset by positive IJDI-Scan [PART 2 OF 2]. The subset is shown for each value of $\lambda \in \{0,0.3,1,3,10\}$ and each number of previous corrections (0, 1, or 2). N/A represents that no subset was found by IJDI-Scan for $\lambda \ge 3$.}
  \label{table:compas-subset-pos-part2}
\end{table*}

\clearpage
%%%%%%%%%%%%%%%%%%%%%%

\begin{table*}
  \begin{center}
    \begin{tabular}{cc|c|ccccc} 
     $\lambda$ & Corrections & Score & Sex & Under 25 & Job & Housing & Savings \\
     \hline
     0 & 0 & 53.5 & all & all & all & all & [Lit., Mod.] \\
     0 & 1 & 27.8 & all & all & all & all & all \\
     0 & 2 & 21.3 & Female & True & {[1, 2+]} & Rent & all \\
     0.3 & 0 & 53.5 & all & all & all & [Free, Rent] & all \\
     0.3 & 1 & 27.8 & all & all & all & [Own, Rent]  & all $\setminus$ [N/A, Rich] \\
     0.3 & 2 & 21.3 & all & all & all &  all & all \\
     1 & 0 & 30.7 & all & all & all & [Free, Rent] & [Lit., Mod.] \\ 
     1 & 1 & 17.8 & all & all & [1, 2+] & all & all \\
     1 & 2 & 13.4 & Female & all & all & all & [Lit., Quite Rich] \\
     3 & 0 & 15.2 & Female & all & all & all & all $\setminus$ [N/A, Rich] \\
     3 & 1 & 7.9 & all & all & all & all & all \\ 
     3 & 2 & 4.5 & all & all & all & Own & all \\
     10 & 0 & 2.7 & all & all & all & Own & all \\
     10 & 1 & 14.5 & Female & all & all & Free & all \\
     10 & 2 & 5.8 & all & all & 2+ & all & Mod. \\
    \end{tabular}
  \end{center}
  \caption[Descriptions of biased negative subgroups for German Credit (Part 1).]{Descriptions of the first three biased subgroups (with significantly high FPR after adjusting for base rates) detected and mitigated in the German credit dataset by negative IJDI-Scan [PART 1 OF 2]. The subset is shown for each value of $\lambda \in \{0,0.3,1,3,10\}$ and each number of previous corrections (0, 1, or 2). Abbreviations: ``Lit." = ``Little", ``Mod." = ``Moderate".}
  \label{table:german-credit-subset-neg-part1}
\end{table*}

%%%%%%%%%%%%%%%%%%%%%%%%

\begin{table*}
  \begin{center}
    \begin{tabular}{cc|cccc} 
     $\lambda$ & Corrections & Checking & Credit Amount & Duration & Purpose \\
     \hline
     0 & 0 & Lit. & all & all $\setminus$ [V. Sh., Sh.] & all \\
     0 & 1 & [Lit., Mod.] & all & [Long, V. Long] & [Bus., Car, Edu.] \\
     0 & 2 & all $\setminus$ [N/A] & all & all & [Bus., Edu., TV] \\
     0.3 & 0 & Lit. & all & all $\setminus$ [V. Sh., Sh.] & [Car, Edu., Eq.] \\
     0.3 & 1 & [Lit., Mod.] & all & [Long, V. Long] & all $\setminus$ [TV, Eq., App.] \\
     0.3 & 2 & all $\setminus$ [N/A] & [Low, V. High] & all $\setminus$ [V. Sh., Sh.] & all $\setminus$ [Rep., Oth., App.] \\
     1 & 0 & Lit. & all & all $\setminus$ [V. Sh., Sh.] & [Car, Edu., Eq.] \\
     1 & 1 & [Lit., Mod.] & all & all $\setminus$ [V. Sh., Sh.] & [Bus., Edu.] \\
     1 & 2 & all $\setminus$ [N/A] & all $\setminus$ [Mod.] & all $\setminus$ [V. Sh., Sh.] & [Car, Edu.] \\
     3 & 0 & [Lit., Mod.] & all & all $\setminus$ [V. Sh., Sh.] & [Bus., Car, Edu., Eq.] \\
     3 & 1 & all & {[Low, V. High]} & all $\setminus$ [V. Sh., Sh.] & [Bus., Edu., Eq., Oth.] \\
     3 & 2 & {[Lit., Mod.]} & all & V. Long & {[Bus., Car]} \\
     10 & 0 & Mod. & all & V. Long & [Bus., Car] \\
     10 & 1 & [Lit., Mod.] & all & all & all \\
     10 & 2 & Lit. & all & all & [Bus., Car] \\
    \end{tabular}
  \end{center}
  \caption[Descriptions of biased negative subgroups for German Credit (Part 2).]{Descriptions of the first three biased subgroups (with significantly high FPR after adjusting for base rates) detected and mitigated in the German credit dataset by negative IJDI-Scan [PART 2 OF 2]. The subset is shown for each value of $\lambda \in \{0,0.3,1,3,10\}$ and each number of previous corrections (0, 1, or 2). Abbreviations: ``V." = ``Very", ``Sh." = ``Short", ``Lit." = ``Little", ``Mod." = ``Moderate", ``Bus." = ``Business", ``Edu." = ``Education", ``TV" = ``Radio/TV", ``Eq." = ``Furniture/Equipment", ``App." = ``Domestic Appliances", ``Rep." = ``Repairs", ``Oth." = ``Vacation/Other".}
  \label{table:german-credit-subset-neg-part2}
\end{table*}

%%%%%%%%%%%%%%%%%%%%%%%
\clearpage
% POSITIVE PART 1
\begin{table*}
  \begin{center}
    \begin{tabular}{cc|c|ccccc} 
     $\lambda$ & Corrections & Score & Sex & Under 25 & Job & Housing & Savings \\
     \hline
     0 & 0 & 37.9 & all & all & all & [Free, Rent] & Lit. \\
     0 & 1 & 23.1 & all & all & all & all & all \\
     0 & 2 & 20.5 & all & all & all & [Own, Rent] & all \\
     0.3 & 0 & 34.1 & all & all & all & [Free, Rent] & Lit. \\
     0.3 & 1 & 21.7 & all & all & all & all & all \\
     0.3 & 2 & 22.4 & all & all & 2+ & [Own, Rent] & Lit. \\
     1 & 0 & 44.3 & all & all & [None, 2+] & all & Lit. \\
     1 & 1 & 21.4 & all & all & [1, 2+] & [Free, Rent] & all \\
     1 & 2 & 19.1 & all & all & all & all & Lit. \\
     3 & 0 & 44.0 & all & all & all & Rent & all \\
     3 & 1 & 63.0 & all & all & [None, 2+] & all & [Lit., Mod.] \\
     3 & 2 & 44.2 & Female & False & all & all & [Lit., Mod.] \\
     10 & 0 & 42.5 & all & True & all & [Free, Rent] & Lit. \\
     10 & 1 & 1.6 & Female & all & all & all & all \\
     10 & 2 & 0.9 & all & all & 1 & all & all \\
    \end{tabular}
  \end{center}
  \caption[Descriptions of biased positive subgroups for German Credit (Part 1).]{Descriptions of the first three biased subgroups (with significantly high TPR after adjusting for base rates) detected and mitigated in the German credit dataset by positive IJDI-Scan [PART 1 OF 2]. The subset is shown for each value of $\lambda \in \{0,0.3,1,3,10\}$ and each number of previous corrections (0, 1, or 2). Abbreviation: ``Lit." = ``Little", ``Mod." = ``Moderate".}
  \label{table:german-credit-subset-pos-part1}
\end{table*}

%%%%%%%%%%%%%%%%%%%%%%%%%
% POSITIVE PART 2

\begin{table*}
  \begin{center}
    \begin{tabular}{cc|cccc} 
     $\lambda$ & Corrections & Checking & Credit Amount & Duration & Purpose  \\
     \hline
     0 & 0 & [Lit., Mod.] & all & [Long, V. Long] & all \\
     0 & 1 & [Lit., Mod.] & [Low, V. High] & all $\setminus$ [V. Sh., Sh.] & all $\setminus$ [TV] \\
     0 & 2 & [Lit., Mod.] & all & V. Long & [Bus., Car, App., Eq.] \\
     0.3 & 0 & [Lit., Mod.] & all & [Long, V. Long] & all \\
     0.3 & 1 & [Lit., Mod.] & [Low, V. High] & all $\setminus$ [V. Sh., Sh.] & all $\setminus$ [TV] \\
     0.3 & 2 & Lit. & all & [Long, V. Long] & [Bus., Car, App., Edu.] \\
     1 & 0 & Lit. & all & [Long, V. Long] & [Bus., Car, App., Edu.] \\
     1 & 1 & [Lit., Mod.] & all & all $\setminus$ [V. Sh., Sh.] & all \\
     1 & 2 & [Lit., Mod.] & all & V. Long & [Bus., Car, Edu., Oth.] \\
     3 & 0 & Lit. & all & [Mod., Long] & all $\setminus$ [Rep., Oth., App.] \\
     3 & 1 & [Lit., Mod.] & all & [Long, V. Long] & [Car, App., Edu., Oth.] \\
     3 & 2 & Lit. & all $\setminus$ High & all & [App., Edu., Eq., Rep.] \\
     10 & 0 & Lit. & all & Long & all \\
     10 & 1 & Lit. & Mod. & all & [Car, Eq.] \\
     10 & 2 & all & V. High & all & all \\
    \end{tabular}
  \end{center}
  \caption[Descriptions of biased positive subgroups for German Credit (Part 2).]{Descriptions of the first three biased subgroups (with significantly high TPR after adjusting for base rates) detected and mitigated in the German credit dataset by positive IJDI-Scan [PART 2 OF 2]. The subset is shown for each value of $\lambda \in \{0,0.3,1,3,10\}$ and each number of previous corrections (0, 1, or 2). Abbreviations: ``V." = ``Very", ``Sh." = ``Short", ``Lit." = ``Little", ``Mod." = ``Moderate", ``Bus." = ``Business", ``Edu." = ``Education", ``TV" = ``Radio/TV", ``Eq." = ``Furniture/Equipment", ``App." = ``Domestic Appliances", ``Rep." = ``Repairs", ``Oth." = ``Vacation/Other".}
  \label{table:german-credit-subset-pos-part2}
\end{table*}
\clearpage

\section{Proving the Effectiveness of Mitigation Approach 3}
\label{sec:randomization}
In Section~\ref{sec:mit-3}, we suggest a possible approach to mitigating IJDI which involves drawing thresholds $\theta_i$ uniformly on $[\theta-\Delta, \theta+\Delta]$, $\Delta \le 0.5$. In this section, we show that  this randomization approach guarantees that no IJDI is present for any $\lambda \ge \frac{1}{2\Delta}$.

Assume two subgroups $A$ and $B$, where the true probability $p_i = \Pr(y_i=1)$ is equal to $p_A$ for all $i \in A$ and $p_B$ for all $i \in B$, $p_A > p_B$. Further, assume a perfectly calibrated predictive model, $\hat p_{i,0} = p_i$, and a sharp threshold for binary recommendations, $\hat p_{i,b} = \mathbf{1}\{\hat p_{i,0} > \theta_i \}$, for all $i$. If $\theta_i$ is equal to a constant $\theta$ for all $i$, then IJDI may exist for any $\lambda \ge 0$.  For example, if $p_A = \theta+\epsilon$ and $p_B = \theta-\epsilon$ for positive constant $\epsilon \approx 0$, $FPR(A)-FPR(B) = 1$, while $p_A - p_B = 2\epsilon \approx 0$, and thus we may have $FPR(A)-FPR(B) > \lambda(p_A - p_B)$ for arbitrarily large values of $\lambda$.

However, if $\theta_i$ is chosen uniformly at random on an interval $[a,b]$, where $0 \le a < b \le 1$, then it can be shown that this randomization approach guarantees $FPR(A) - FPR(B) \le \lambda (p_A - p_B)$, and thus no IJDI is present, for any $\lambda \ge \frac{1}{b-a}$.  To see this, consider the probability density function $f(\theta)$, where $\theta_i \:{\sim}\: f(\theta)$ for all $i$.  For the uniform case, we know that $f(\theta) = \frac{1}{b-a}$ for $a \le \theta \le b$, and 0 otherwise. Then:
\begin{align*}
FPR(A) - FPR(B) &= \Pr(\theta_i < p_A) - \Pr(\theta_i < p_B) \\
&= \Pr(p_B \le \theta_i < p_A) \\
&= \int_{\theta = p_B}^{p_A} f(\theta) d\theta \\
&\le \frac{p_A-p_B}{b-a}.
\end{align*}
Finally, if $\lambda \ge \frac{1}{b-a}$, it follows that $FPR(A) - FPR(B) \le \frac{p_A - p_B}{b-a} \le \lambda(p_A - p_B)$, and no IJDI is present for that value of $\lambda$.

\section{Limitations of the Work}
\label{sec:limitations}

Perhaps the most important limitation of our IJDI fairness criterion is that whether a given FPR or TPR disparity is considered ``sufficiently'' or ``insufficiently'' justified is critically dependent on the user's choice of $\lambda_N$ and $\lambda_P$.  These parameters significantly impact how base rates (utilities) factor into the scan and trade off with error rate balance, and this tradeoff might differ for different application domains.  While we have provided some guidance on preference elicitation (Appendix~\ref{sec:eliciting}), this question should be further explored, incorporating an understanding of human heuristics and biases, to discover whether users can choose reasonable $\lambda$ values.  Further, an important question is who gets to specify these critical parameters. Is the person making this decision adequately representative of the people whom their decisions will be affecting?  One approach to addressing this limitation might be participatory approaches to obtaining these parameters, where preferences are elicited from many different constituent groups, particularly focusing on subpopulations who are affected by the deployment of a predictive model but might not otherwise get a say in how the model is used.  A simple alternative would be to run IJDI-Scan multiple times, over a wide array of $\lambda_N$ and $\lambda_P$ values, to audit the fairness of a predictive model, and let different people reach their own decisions about whether the classifier being audited is fair based on their own desired tradeoffs between utility and error rate balance.  This simpler approach might be inadequate for making concrete policy decisions, e.g., whether error rate disparities are so large as to recommend against the use of a predictive model, but we would argue that such decisions would require a broader context than simply whether or not significant IJDI is present in any case.

Two technical limitations of the work are (i) the assumption that all covariates are discrete rather than real-valued, and (ii) the need to separately assess FPR and TPR disparities by running IJDI-Scan for the negative individuals ($D_N$) and positive individuals ($D_P$) of dataset $D$.  To address the first limitation, we note that real-valued attributes can be 
discretized as a preprocessing step, using either the observed distribution or domain knowledge, as we have done with the COMPAS and German Credit datasets.  Nevertheless, discretization of real-valued attributes might introduce additional bias into the data.  To address the second limitation, future work could include developing a ``combined'' IJDI-Scan that returns the single subgroup with the largest combined discrepancy in IJDI for both FPR \emph{and} TPR. Such a combined scan could take both $\lambda_N$ and $\lambda_P$ as parameters, and a specific domain's sensitivity to $\Delta_{FPR}$ and $\Delta_{TPR}$ would determine whether the negative or positive IJDI criterion dominates. 

As noted in Section~\ref{sec:ijdi}, the true probabilities $p_i$ for each individual are often not known, and can only be estimated from the observed data. While our experiments in Section~\ref{sec:exp-2-learned} demonstrate that IJDI-Scan can achieve comparable performance with learned and true probabilities, assuming a sufficiently flexible model and sufficient data, poor estimates of $p_i$ (e.g., because of a misspecified model or limited data) could introduce biases into the IJDI-Scan results.

Lastly, we note the complexity of the topic of fairness in criminal justice and specifically, issues related to the use of COMPAS as a benchmark dataset. Previous work have questioned the use of pretrial risk assessment instrument (RAI) datasets for evaluating algorithms without adequate regard for the complexity of the domain and have pointed out measurement errors and biases\footnote{For example, see Bao et al., ``It's COMPASlicated: The Messy Relationship between RAI Datasets and Algorithmic Fairness Benchmarks'', arXiv:2106.05498, 2021.}. In this work, we have not used the COMPAS risk assessments as ground truth but instead critically evaluate this tool, demonstrating both (known) racial biases and (previously unknown) intersectional gender and age biases. Secondly, where needed in the experiments, we have generated synthetic predictions and outcomes to be able to compare our empirical results to known ground-truth values, rather than using the COMPAS predictions.  More generally, we are certainly not arguing for the use of algorithmic decision support tools in pre-trial risk assessment, but rather, we present an auditing tool that could be used to identify biases in either algorithmic or human decisions.  Finally, we have sufficiently demonstrated the usefulness of IJDI-Scan in assessing error rate balance for algorithmic decision support tools in contexts beyond criminal justice.  As noted above, a finding of whether significant IJDI is present is not sufficient to draw prescriptive conclusions about whether the harms of 
an algorithmic tool outweigh its benefits, but provides one piece of evidence that can be considered along with the broader context to assist in these decisions.

\section{Reproducibility Checklist}
\label{sec:reproducibility}
\subsection{Code} All relevant code can be found in the public repository: \verb|https://github.com/| \verb|neil-menghani/ijdi|.

\subsection{Hardware} For this paper, experiments were run on Windows 11 Pro, 16 GB RAM. However, the code can be run from Windows, Mac OS, or Linux (8+ GB RAM recommended).

\subsection{Software} Python 3.7.13 and Anaconda 4.13.0. See \verb|environment.yml| for a detailed list of required libraries.

\subsection{Randomness}
To replicate the results from the paper, set random seed to 100 for each experiment. However, given enough iterations the results from each experiment will converge to the same approximate value.

\subsection{Instructions for running experiments}
Functions used by the experiments can be found in the \verb|scripts| folder, including IJDI-Scan (Algorithm~\ref{alg:algorithm}) defined in \verb|scan.py| as \verb|run_ijdi_scan|. For COMPAS experiments, the dataset \verb|compas.csv| and iPython notebooks for pre-processing the data, running experiments, and generating plots can be found in the \verb|compas| directory. For German Credit experiments, the dataset \verb|german_credit.csv| and iPython notebooks can be found in the \verb|german_credit| directory. For a more detailed list of relevant notebooks, see \verb|README.md|.

\end{document}